\pdfoutput=1
\documentclass[11pt]{article}

\usepackage{acl}

\usepackage{times}
\usepackage{latexsym}

\usepackage[T1]{fontenc}
\usepackage[utf8]{inputenc}

\usepackage{microtype}

\usepackage{inconsolata}

\usepackage{times}
\usepackage{array}
\usepackage{amsmath}
\usepackage{siunitx}
\usepackage{latexsym}
\usepackage{booktabs}
\usepackage{longtable}
\usepackage{multirow}
\usepackage{graphicx}
\usepackage{graphics}
\usepackage{caption}
\usepackage{subcaption}
\usepackage{xcolor}
\usepackage{multicol}
\usepackage{colortbl}
\usepackage{url}
\usepackage{xurl}
\usepackage{tabularx}
\usepackage{algorithm}
\usepackage{algpseudocode}
\usepackage{soul}
\usepackage{bm}
\usepackage{tcolorbox}
\usepackage{amsmath}
\usepackage{bbm}

\title{\textsc{SpeciaLex}: A Benchmark for In-Context Specialized\\Lexicon Learning}

\author{Joseph Marvin Imperial$^{\Omega,\Lambda}$~\;~Harish Tayyar Madabushi$^{\Lambda}$ 
\\$^{\Lambda}$University of Bath, UK\\ $^{\Omega}$National University, Philippines
\\\texttt{\href{mailto:jmri20@bath.ac.uk}{jmri20@bath.ac.uk}}~\;~
\texttt{\href{mailto:htm430@bath.ac.uk}{htm43@bath.ac.uk}}
}

\begin{document}
\maketitle
\begin{abstract}

Specialized lexicons are collections of words with associated constraints such as special definitions, specific roles, and intended target audiences. These constraints are necessary for content generation and documentation tasks (e.g., writing technical manuals or children's reading materials), where the goal is to reduce the ambiguity of text content and increase its overall readability for a specific group of audience. Understanding \textit{how} large language models can capture these constraints can help researchers build better, more impactful tools for wider use beyond the NLP community. Towards this end, we introduce \textsc{SpeciaLex}, a benchmark for evaluating a language model's ability to follow specialized lexicon-based constraints across $18$ diverse subtasks with $1,785$ test instances covering core tasks of \textsc{Checking}, \textsc{Identification}, \textsc{Rewriting}, and \textsc{Open Generation}. We present an empirical evaluation of $15$ open and closed-source LLMs and discuss insights on how factors such as model scale, openness, setup, and recency affect performance upon evaluating with the benchmark.\footnote{The task datasets and evaluation code can be found at: \url{https://github.com/imperialite/specialex/}.}
\end{abstract}

\section{Introduction}
The adoption of large language models (LLMs) for domains beyond computing and AI has been more evident in recent years, particularly with the release of publicly accessible chat interfaces such as ChatGPT. This widespread use from various multidisciplinary communities can be primarily attributed to modern LLMs' capabilities to learn patterns from just a few examples during inference---\textit{in-context learning (ICL)}---combined with the use of modern architectures and massive and diverse datasets to train them to follow complex instructions \cite{wei2022chain,chung2022scaling,brown2020language}. With in-context learning, LLMs can be treated as task-agnostic systems and can do virtually any text-related task, including open-ended generation and structured prediction, just by being conditioned to provide completions for prompts given task-specific demonstrations \cite{brown2020language,radford2019language,radford2018improving}.

%%
%% - SpeciaLex TYPOLOGY
%% 
%trim={left bottom right top}
\begin{figure}[!t]
    \centering
    \includegraphics[width=.50\textwidth,trim={17cm 4.5cm 17cm 2cm}, clip]{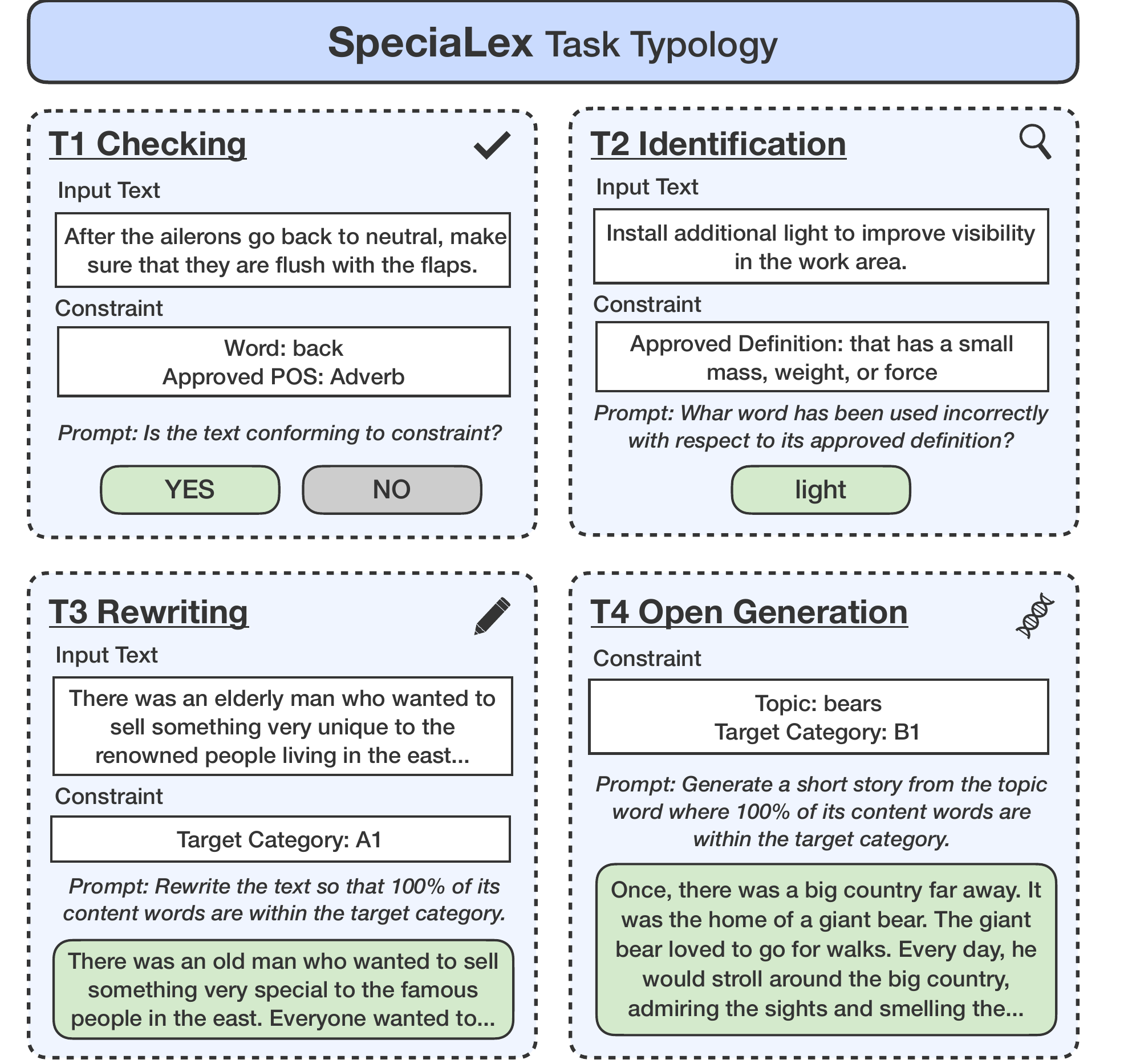}
    \caption{An overview of the task coverage of \textsc{SpeciaLex}. The examples shown for \textsc{Checking} and \textsc{Identification} use constraints from the Simple Technical English (STE) lexicon for technical writing in engineering, while the examples for \textsc{Rewriting} and \textsc{Open Generation} are from the Oxford 5000 lexicon for content generation in education.}
    \label{fig:specialex-typology}
\end{figure}

%%
%% - SpeciaLex VIS 
%% 
%trim={left bottom right top}
%\begin{figure}[!t]
%    \centering
%    \includegraphics[width=.45\textwidth,trim={9cm 7cm 34cm 3cm}, clip]{figs/SpeciaLex-vis.pdf}
%    \caption{An overview of \textsc{SpeciaLex} for evaluating capabilities of state-of-the-art LLMs for capturing specialized lexicon-based constraints (specific roles, {special definitions}, and {target audience}) tested across four tasks (Checking, {identification}, {rewriting}, {open generation}).}
%    \label{fig:SpeciaLex-vis}
%\end{figure}

One particular point of interest in the wider adoption of LLMs is evaluating how they can capture lexicon-based constraints for generating text content across different domains. For example, in education, a teacher who knows how to masterfully use an LLM (e.g., ChatGPT) to generate classroom-ready reading materials on the fly can accommodate students' various interests in reading \cite{kasneci2023chatgpt}, such as prompting the LLM with preferred topics for stories and custom character roles. However, if used this way, the LLM should learn constraints such as knowing what specific words are readable by a target audience (e.g., ages 10-11). These special words are often found on specially curated lexicons such as the Oxford 5000 Wordlist\footnote{\url{https://www.oxfordlearnersdictionaries.com/wordlists/}}. In technical writing, on the other hand, an LLM should learn to capture customized word definition constraints as mandated by existing guidelines and standards to avoid producing ambiguous texts. For example, as per Simplified Technical English (STE)\footnote{\url{https://www.asd-ste100.org/}} guidelines, the word \textit{glue} cannot be used as a verb to mean \textit{stick together}; the appropriate word for this is \textit{bond} or \textit{attach}. 

Understanding how current LLMs capture fine-grained constraints from specialized lexicons across domains opens a number of opportunities for improving their ability to follow instructions at a very fine level, particularly through in-context learning. However, the main gap here is that there are currently no comprehensive evaluation studies or benchmarks to guide researchers in learning more about the performance and limitations of modern LLMs on content generation tasks requiring compliance with said constraints. 

In this study, we fill the gap by introducing \textsc{SpeciaLex}, a comprehensive benchmark suite composed of $18$ diverse tasks to evaluate the capabilities of LLMs in capturing lexicon-based constraints such as special roles or part-of-speech, special word definitions, and target audiences. We provide an in-depth comparison of $15$ state-of-the-art LLMs as baselines and release extendable \textsc{SpeciaLex} subtask data comprising $1,785$ test instances. We devised four core task variations spanning \textsc{Checking}, \textsc{Identification}, \textsc{Rewriting}, and \textsc{Open Generation}. Implementation-wise, we structured \textsc{SpeciaLex} to focus on using \textit{in-context learning} for all tasks as this emulates the most common way for lay people and users to interact with LLMs through carefully structured prompts with examples or demonstrations.

By evaluating a diverse set of commercial and open LLMs in terms of task performance, scale, and openness, \textsc{SpeciaLex} serves as a valuable reference and guide for interdisciplinary researchers who require the use of capable LLMs but are on a limited computing budget or are concerned only with performance on specific constraints. Moreover, by following design principles from established open LLM benchmarks such as \textsc{LegalBench} \cite{guha2024legalbench}, the research community can extend and build upon \textsc{SpeciaLex} by contributing new tasks and specialized lexicons from other domains to expand the evaluation of LLMs in this direction.

%\textcolor{red}{Move this as a footnote in the abstract -->}For a wider community adoption of this benchmark, we will release \textsc{SpeciaLex} including all research artifacts associated with its development (code, data, model generations) upon publication of this paper.
%For example, \textit{model A performs well on capturing constraint B but not on A, C, and D. However, constraint B is the only requirement for domain X. Therefore, model A is a good fit for domain X}.

%Thus, the main contributions of this work are as follows:
%\begin{enumerate}
%    \item We release \textsc{SpeciaLex}, a comprehensive benchmark suite composed of 18 diverse tasks to evaluate the capabilities of LLMs to check, identify, rewrite, and generate text based on lexicon constraints covering special roles, special definitions, and target audiences.
%    \item We release \textsc{SpeciaLex}-ready datasets for the domains of education (CEFR) and English technical writing (ASD-STE) composed of 1700+ test instances as reference for researchers to build upon \textsc{SpeciaLex} or use domain-specific corpora as an extension.
%    \item We provide an in-depth comparison of 15 state-of-the-art LLMs as baselines across increasing scales and openness with \textsc{SpeciaLex}.
%\end{enumerate}

\section{Related Work}

\noindent \textbf{Benchmarks for Content Generation}. Parallel to its widespread adoption, the rise of benchmark studies has also gained significant traction from the LLM community. For generative tasks, existing works have explored evaluating general aspects such as factuality \cite{muhlgay-etal-2024-generating}, model hallucinations \cite{li-etal-2023-halueval}, safety and toxicity \cite{rottger2023xstest,hartvigsen-etal-2022-toxigen,gehman-etal-2020-realtoxicityprompts}, low-resource language and multilingual capabilities \cite{chen-etal-2022-mtg,liang-etal-2020-xglue}, and surface-level properties and lexical constraints \cite{kew-etal-2023-bless,sun-etal-2023-evaluating,gehrmann-etal-2021-gem} to name a few. To our knowledge, no existing benchmark has yet to consider evaluating LLMs for capturing special definitions, specific roles or part-of-speech, and knowledge of recognizable words of target audiences, which \textsc{SpeciaLex} aims to fulfill.

\noindent \textbf{Augmenting Lexicons and Dictionaries to LLMs}. The use of lexicons and dictionaries has served as an additional knowledge base for LLMs across a number of tasks. \citet{he-yiu-2022-controllable} used the Oxford dictionary to finetune BART models to generate appropriate sentence examples based on words. \citet{yu-etal-2022-dict} used dictionary definitions of rare words to improve the pre-training of LLMs. Similarly, \citet{wu-etal-2022-dictionary} also used specialized lexicons to improve the contrastive learning objective of pertaining BERT and RoBERTa models for tasks such as abusive language detection and sentiment analysis. Our use of lexicons for \textsc{SpeciaLex} serves as a reference of constraint for LLMs for content generation tasks. Moreover, while all the previous works cited make use of extra training via finetuning to make their models task-specific, \textsc{SpeciaLex} focuses on capturing constraints purely by in-context learning while preserving the evaluated models' ability to perform across general tasks.

\noindent \textbf{Domain Adaptation of LLMs}. Researchers from interdisciplinary fields are working with the NLP community to evaluate the domain-specific capabilities of LLMs. A few of these collaborations include notable works such as \textsc{LegalBench} \cite{guha2024legalbench} with $162$ tasks for legal reasoning, \textsc{ChemLLMBench} \cite{guo2023can} with $8$ tasks for understanding, explaining, and prediction tasks in practical chemistry, \textsc{RAFT} \cite{alex2021raft} with $11$ multidisciplinary tasks, and \textsc{PubMedQA} \cite{jin-etal-2019-pubmedqa}, \textsc{MedMCQA} \cite{pmlr-v174-pal22a}, and \textsc{MedBench} \cite{cai2024medbench} for biomedical question answering. \textsc{SpeciaLex} draws similar motivation with \textsc{LegalBench} \cite{guha2024legalbench}, \textsc{RAFT} \cite{alex2021raft}, and \textsc{ChemLLMBench} \cite{guo2023can} in terms of benchmark typology and evaluation method via in-context learning, which is further expanded in the succeeding sections.\\

\definecolor{constraint_color}{RGB}{224, 255, 226}
\definecolor{task_color}{RGB}{215, 236, 252}
%\sethlcolor{constraint_color}

\section{\textsc{SpeciaLex}: A Benchmark for In-Context Specialized Lexicon Learning}
\label{sec:SpeciaLex-definition}

We build \textsc{SpeciaLex} as a general benchmark and reference for evaluating LLMs to capture lexicon-based constraints through in-context learning. We discuss the task typology and recognized lexicon-based constraints of \textsc{SpeciaLex} as seen in Figure~\ref{fig:specialex-typology}.

\subsection{Constraint Types}
We select three general lexicon-based constraint types for \textsc{SpeciaLex} as the reference for controlling the generation of text content from LLMs. The selection of these constraints has been derived from consultations with domain experts (further discussed in Section~\ref{sec:SpeciaLex-tasks}) and from surveying the overlap of constraints from existing works on dictionary-based augmentation with LLMs \cite{he-yiu-2022-controllable} and controllable text generation \cite{sun-etal-2023-evaluating,zhou2023controlled}. We describe the conditions of each lexicon-based constraint below: \\

\noindent \textbf{\textsc{\colorbox{constraint_color}{C1 - Specific Roles}}} describes the constraint that restricts a word from a lexicon from having multiple roles via part-of-speech (POS) information in a text and recommends an alternative word with a specific POS. For example, the word \textit{brush} can only be used as a noun referring to the cleaning material and not as a verb referring to \textit{brushed} or \textit{brushing} and should be treated as the replacement word for unapproved words such as \textit{scrub}. Evaluation-wise, an LLM must be able to generate a text where a given word is replaced with its alternative and its approved POS. This constraint is particularly prevalent in technical writing guidelines such as Simple Technical English (STE) for developing manuals to reduce context ambiguity \cite{knezevic2015improving}. \\

\noindent \textbf{\textsc{\colorbox{constraint_color}{C2 - Special Definition}}} describes the constraint that a word must be used according to its special domain-specific definition. Similar to \textsc{Specific Roles}, this helps significantly reduce ambiguity in writing given that the common English language uses homonyms\footnote{Words with two or more meanings.}. For example, in Simple Technical English (STE), the word \textit{close} in a sentence should only mean \textit{blocking of entrance} and not having \textit{two materials near each other}. Evaluation-wise, a model must ensure that the special definition of a word is preserved in the text.\\  

\noindent \textbf{\textsc{\colorbox{constraint_color}{C3 - Target Audience}}} describes the constraint that target audiences or readers are associated with specific groups of words that domain experts think they can easily read. Evaluation-wise, an LLM must be able to maximize the use of readable words appropriate for a target audience for generating content. An example constraint resource for this is the Oxford 5000 lexicon, containing sets of words for each increasing level in the CEFR scale (A1, A2, B1, B2, and C1) curated by experts in language assessment. In \textsc{SpeciaLex}, we explore two levels of conformity $c$ to the resource lexicons for the target audience: full ($c = 1.0$) and minimal ($c = 0.95$). We draw support from empirical studies in reading such as by \citet{laufer198925} and \citet{hsueh2000unknown}, which states that a reading material must have \textit{at least} $95\%$ of the content words readable by a learner to ensure effective comprehension of the text. Through \textsc{SpeciaLex}, researchers from other domains can explore setting different levels of conformity based on their theoretical grounding. \\

%%
%% TASK OVERVIEW
%%
\begin{table}[!t]
\centering
\small
\renewcommand{\arraystretch}{1.0}
\begin{tabular}{@{}lcccc@{}}
\toprule
\multicolumn{1}{c}{\multirow{2}{*}{\bf Tasks}} & \multicolumn{4}{c}{\bf Constraints} \\ \cmidrule{2-5}
\multicolumn{1}{c}{} &
  \begin{tabular}[c]{@{}c@{}}\bf \textsc{C1}\end{tabular} &
  \begin{tabular}[c]{@{}c@{}}\bf \textsc{C2}\end{tabular} &
  \begin{tabular}[c]{@{}c@{}}\bf \textsc{C3}\end{tabular} &
  \bf \textsc{(\textsc{C1+C2})} \\ \midrule
 \textsc{Checking}                              & 72     & 64     & 115   & -     \\
 \textsc{Identification}                             & 77     & 69     & 108   & -     \\
 \textsc{Rewriting}                                  & 300    & 82     & 106   & 67    \\
 \textsc{Open Generation}                            & 175    & 175    & 200   & 175  \\
\bottomrule
\end{tabular}
\caption{A summary of breakdown of test instances for each core task and constraint covered by \textsc{SpeciaLex}. A more complete version with the extensive definitions can be found in Appendix~\ref{app:appendix}.}
\label{tab:SpeciaLex-task-overview}
\end{table}

\subsection{Task Typology}
For each task $T$, we define a prompt $p$, which describes the official task instruction as an input to the LLM and a set of task-specific demonstrations $d_{n}$ conforming to a constraint $c$. We set $n = 5$ as the minimum number of in-context learning examples similar with existing benchmarks such as \textsc{LegalBench} \cite{guha2024legalbench} and \textsc{RAFT} \cite{alex2021raft}. We describe the setup for each task below: \\

\noindent \textbf{\textsc{\colorbox{task_color}{T1 - Checking}}} involves validation of a given input text whether to conforms to a specified constraint. As a validation task, the constraint can only be one of the three recognized \textsc{SpeciaLex} constraints. The outputs for \textsc{Checking} tasks are binary \textsc{YES} or \textsc{NO}. There are a total of $4$ \textsc{Checking} tasks in \textsc{SpeciaLex}. \\

\noindent \textbf{\textsc{\colorbox{task_color}{T2 - Identification}}} is another validation-type task that involves listing (non)conformity of an input text from a given task and lexicon-based constraint. The variation of \textsc{Identification} spans recognizing what word or set of words violate specific roles, special definitions, or target audience assigned by recognized constraints as well as identifying the most appropriate correct target audience. There are a total of $4$ \textsc{Identification} tasks in \textsc{SpeciaLex}.\\

\noindent \textbf{\textsc{\colorbox{task_color}{T3 - Rewriting}}} involves reconstructing an input text that violates a given lexicon-based constraint into a correct version which will be evaluated accordingly. We consider \textsc{Rewriting} as a semi-open generation task since the output is no longer structured like \textsc{Checking} or \textsc{Identification}, but the LLM still has a reference to the incorrect version and in-context demonstrations as guidance. There are a total of $5$ \textsc{Rewriting} tasks in \textsc{SpeciaLex}.\\

\noindent \textbf{\textsc{\colorbox{task_color}{T4 - Open Generation}}} is a full open-ended generative task that requires the LLM to generate a constraint-compliant output on-the-fly from the input text and task-specific demonstrations. Moreover, unlike \textsc{Rewriting}, each \textsc{Open Generation} task instance has no reference to an incorrect version and only the word and its associated constraint it needs to generate with, which makes this task more challenging. There are a total of $5$ \textsc{Open Generation} tasks in \textsc{SpeciaLex}.

\section{\textsc{SpeciaLex} Task Construction Process}
\label{sec:SpeciaLex-tasks}

This section provides an overview of the construction process we followed for building and evaluating tasks for \textsc{SpeciaLex} with resources provided by experts.

\subsection{Collaborative Element} 
Throughout this study's development, we collaborated with two domain expert representatives from the Simplified Technical English Maintenance Group (STEMG) and one from the Common European Framework of Reference for Languages (CEFR)\footnote{\url{https://www.coe.int/en/web/common-european-framework-reference-languages}}. We covered discussions for the acquisition of shareable machine-readable corpora, the conduct of periodical discussions of experiment results, and validation of automatic metrics used for \textsc{SpeciaLex} described in the succeeding subsections. With this, we consider \textsc{SpeciaLex} as an LLM benchmark where domain experts have significantly contributed to its design and development.

\subsection{Specialized Lexicon Data}
For constructing the test cases in \textsc{SpeciaLex}, we use globally recognized specialized lexicons in English, both used in technical writing and language assessment described below, to capture the three core constraints described in Section~\ref{sec:SpeciaLex-definition}. 
Note that these lexicons do not require any additional expert annotations as they are already off-the-shelf resources packaged as expert-developed datasets. Additional information can be found in Appendix~\ref{app:datasets}. \\

\noindent \textbf{Simple Technical English Lexicon (STE)} is an international industry-standard specification of controlled language used for simpler and clearer English technical documentation developed by the European Association of Aerospace Industries (AECMA). Previously exclusively used within aerospace engineering, STE has been adopted in many fields, including education, defense, and maintenance, and used across tasks such as machine translation and simplification \cite{kuhn-2014-survey,zambrini2023subject}. STE has a lexicon component that contains $1,259$ words with associated alternative words and part-of-speech information and $939$ with special definitions. These constraints aim to reduce ambiguity and ensure that the text can be easily understood by non-native English speakers. We use the lexicon of STE Issue 7 (released 2017) to manually construct test instances for the tasks classified evaluating \textsc{Specific Roles} and \textsc{Special Definition} constraints for \textsc{SpeciaLex}.\\

%It has since been adopted in many other fields outside the aerospace, defense, and maintenance domains for its clear, consistent, and comprehensive nature. 

\noindent \textbf{Oxford 5000 Lexicon} is an expanded open-source compilation of English words distributed across the associated levels in the Common European Framework of Reference for Languages (CEFR) Framework published by the Oxford University Press. This resource is derived from the Oxford English Dictionary and is widely adopted by CEFR educators. It also guides beginner and advanced learners on what words they should know at each specific CEFR level (from A1 to C1). We use the expanded version with $5,335$ words and their associated CEFR levels to manually construct the test cases for evaluating the \textsc{Target Audience} constraint for \textsc{SpeciaLex}. \\

\subsection{Prompt Construction}
We followed the prompt construction process observed by \textsc{LegalBench} \cite{guha2024legalbench} where, for each subtask, a base prompt is used containing $5$ random gold-standard demonstrations serving as in-context examples and a test file containing the manually constructed test instances with respect to the specific constraint and core task being evaluated by the subtask (e.g., \textsc{Checking} with \textsc{Specific Roles} as visualized in Figure~\ref{fig:specialex-typology}). Each instance in the test file is appended to the base prompt for prompting an LLM to capture its output, which will then be evaluated with a task and constraint-appropriate method. Additional information and actual prompt templates can be found in Appendix~\ref{app:additional_task_infor} and ~\ref{app:task_prompt_templates}.

\subsection{Evaluation}
Our selection of automatic evaluation methods is based on discussions with domain experts and references to previous works. Additional information can be found in Appendix~\ref{app:evaluation}.

Structured prediction and binary classification tasks from \textsc{Checking} and \textsc{Identification} are evaluated using exact-match accuracy as done in other LLM benchmarks \cite{guha2024legalbench,liang2023holistic,alex2021raft}. For \textsc{Rewriting} and \textsc{Open Generation} tasks requiring a model to produce texts conforming to specific roles, special definitions, or words for a target audience, we use varying tools for resolving alignment. For conformity of a word based on a specific role through POS, we use Spacy\footnote{\url{https://spacy.io/api/tagger}} implementation of a POS classifier for identifying the POS information of a target word. For judging whether a word has been used according to its approved definition, we use GPT-4 as a judge. Existing LLM benchmarks and chatbot arenas have used GPT-4 as a judge for its high performance across general and semantic-based tasks, and results have shown a significantly high level of agreement with human experts \cite{zheng2024judging,asai2023self}. For assessing texts based on a target audience, we developed a simple lexicon-matching script that sums the total unique content words (nouns, adjectives, adverbs, verbs) recognized by the target category (e.g., A2) and divided by the total words of the text. Thus, closer values to $1.0$ are better, entailing higher density of words recognized by the target audience.
%Our implementation of these evaluation metrics is normalized to produce a score between $[0.0 - 1.0]$ for uniform comparison.

\subsection{Benchmark Statistics}
Upon completion of the construction process, \textsc{SpeciaLex} contains a total of $1,785$ test instances distributed across $18$ subtasks from the $4$ core task category as reported in Table~\ref{tab:SpeciaLex-task-overview} and in Table~\ref{tab:SpeciaLex-full-table}. Subtasks contain test instances with a minimum of $53$ and a maximum of $300$ (average $99$). We note that these numbers are closely comparable to existing domain-adapted recent LLM benchmarks, including \textsc{LegalBench} \cite{guha2024legalbench} and \textsc{RAFT} \cite{alex2021raft} where the minimum number of tests instances are also set to 50.

%%
%% TABLE - STE TASKS
%%
\newcolumntype{C}[1]{>{\centering\arraybackslash}p{#1}}
\newcolumntype{L}[1]{>{\raggedright\arraybackslash}p{#1}}
\setlength{\tabcolsep}{5pt}
\begin{table*}[!t]
\centering
\small
\renewcommand{\arraystretch}{1.3}
\begin{tabularx}{\textwidth}{@{}L{2.5cm}C{0.7cm}C{0.7cm}C{0.7cm}C{1.1cm}C{0.7cm}C{0.7cm}C{1.3cm}C{0.7cm}C{0.7cm}C{1.3cm}C{1.1cm}@{}}
\toprule
%\midrule
\multirow{2}{*}{\textbf{LLMs}} & 
  \multicolumn{2}{c}{\bf \textsc{Checking}} &
  \multicolumn{2}{c}{\bf \textsc{Identification}} &
  \multicolumn{3}{c}{\bf \textsc{Rewriting}} &
  \multicolumn{3}{c}{\bf \textsc{Open Generation}} &
  \multirow{2}{*}{$\bm{\mu}$}\\ \cmidrule{2-11}
\multicolumn{1}{c}{} & \bf \textsc{ID1} & \bf \textsc{ID2}   & \bf \textsc{ID3}   & \bf \textsc{ID4}   & \bf \textsc{ID5}   & \bf \textsc{ID6}   & \bf \textsc{ID7} & \bf \textsc{ID8}   & \bf \textsc{ID9} & \bf \textsc{ID10} &\\ \midrule
Gemma-2B             & 0.46 & 0.50 & 0.68 & 0.54 & 0.49 & 0.51 & 0.26    & 0.61 & 0.62 & 0.63    & 0.54 \\
OLMO-1B              & 0.50 & 0.05 & 0.52 & 0.71 & 0.46 & 0.36 & 0.43    & 0.09 & 0.88 & 0.12    & 0.40 \\
BLOOM-1B             & 0.50 & 0.50 & 0.74 & 0.67 & 0.58 & 0.42 & \underline{0.51}    & 0.23 & 0.67 & 0.15    & 0.50 \\ \midrule
Llama3-8B            & 0.56 & 0.81 & 0.74 & 0.86 & 0.10 & \underline{0.63} & 0.17    & 0.03 & 0.32 & 0.07    & 0.42 \\
Mistral-7B           & 0.53 & 0.72 & 0.49 & 0.57 & \underline{0.70} & 0.48 & 0.43    & 0.87 & 0.80 & 0.80    & 0.65 \\
Llama2-7B            & 0.50 & 0.50 & 0.43 & 0.71 & \underline{0.70} & \bf 0.67 & 0.41    & 0.83 & 0.73 & 0.78    & 0.64 \\
Llama2-13B           & 0.50 & 0.56 & 0.57 & 0.78 & 0.69 & 0.60 & 0.44    & 0.85 & 0.83 & 0.87    & 0.69 \\
OLMO-7B              & 0.38 & 0.64 & 0.49 & 0.67 & 0.60 & 0.57 & 0.39    & 0.80 & 0.67 & 0.76    & 0.62 \\
Gemma-7B             & 0.53 & 0.34 & 0.66 & 0.71 & 0.69 & 0.51 & 0.47    & 0.80 & 0.77 & 0.80    & 0.64 \\
BLOOM-7B             & 0.50 & 0.50 & 0.44 & 0.59 & 0.66 & \bf 0.67 & \bf 0.69    & 0.57 & 0.34 & 0.25    & 0.52 \\ \midrule
CommandR-105B        & 0.53 & 0.89 & 0.75 & 0.88 & 0.27 & 0.57 & 0.38    & 0.88 & 0.91 & 0.87    & 0.71 \\
Llama2-70B           & 0.53 & 0.13 & 0.55 & 0.88 & 0.27 & 0.59 & 0.48    & 0.85 & 0.87 & 0.87    & 0.61 \\
Llama3-70B           & \underline{0.69} & \underline{0.91} & \bf 0.83 & \underline{0.94} & 0.29 & 0.59 & 0.50    & \bf 0.92 & \underline{0.93} & \underline{0.91}    & {0.76} \\ \midrule
GPT3.5-Turbo         & 0.47 & 0.88 & 0.75 & \bf 0.99 & 0.63 & 0.61 & 0.49    & \underline{0.90} & 0.90 & 0.89    & \underline{0.78} \\
GPT-4o                & \bf 0.89 & \bf 0.94 & \underline{0.82} & 0.93 & \bf 0.75 & 0.62 & 0.48    & \bf 0.92 & \bf 0.97 & \bf 0.94    & \bf\underline{0.82}\\ 
%\midrule
\bottomrule
\end{tabularx}
\caption{Overview of instruction-tuned LLM performances evaluated through \textsc{SpeciaLex} for capturing \textbf{\textsc{C1 (Specific Role)}} and \textbf{\textsc{C2 (Special Definition)}} constraints where test instances were derived from the \textbf{STE lexicon}. Each section division corresponds to the grouped LLMs based on similar scales. Values in bold mean the highest performance, while those underlined are second. Column $\mu$ denotes the mean performance across all subtasks. The underlined value for \textbf{GPT-4o} denotes that it is the overall best-performing model for generating content aligned with the specified constraints. Column names can be referenced through subtask IDs in Table~\ref{tab:SpeciaLex-full-table}.}
\label{tab:SpeciaLex-result-ste}
\end{table*}

\begin{table*}[!htbp]
\centering
\small
\renewcommand{\arraystretch}{1.3}
\begin{tabularx}{\textwidth}{@{}L{2.5cm}C{0.7cm}C{1.5cm}C{1.cm}C{1.5cm}C{0.7cm}C{1.5cm}C{0.7cm}C{1.5cm}C{1.3cm}@{}}
\toprule
%\midrule
\multirow{2}{*}{\textbf{LLMs}} &
  \multicolumn{2}{c}{\bf \textsc{Checking}} &
  \multicolumn{2}{c}{\bf \textsc{Identification}} &
  \multicolumn{2}{c}{\bf \textsc{Rewriting}} &
  \multicolumn{2}{c}{\bf \textsc{Open Generation}} &
  \multirow{2}{*}{$\bm{\mu}$} \\ \cmidrule{2-9}
 &
  \bf \textsc{ID11} &
  \bf \textsc{ID12} &
  \bf \textsc{ID13} &
  \bf \textsc{ID14} &
  \bf \textsc{ID15} &
  \bf \textsc{ID16} &
  \bf \textsc{ID17} &
  \bf \textsc{ID18} &
   \\ \midrule
Gemma-2B      & 0.49 & 0.31 & 0.00 & \underline{0.23} & 0.68 & 0.68 & 0.69 & 0.69 & 0.47 \\
BLOOM-1B      & \underline{0.85} & \underline{0.84} & 0.00 & 0.21 & 0.69 & 0.69 & \underline{0.71} & \bf 0.72 & \underline{0.59} \\ \midrule
Llama3-8B     & \bf 0.96 & \bf 0.94 & 0.00 & \bf 0.30 & 0.68 & 0.69 & 0.69 & 0.70 & \bf \underline{0.62} \\
Mistral-7B    & 0.68 & 0.52 & 0.02 & 0.11 & \underline{0.70} & 0.69 & 0.65 & 0.65 & 0.50 \\
Llama2-7B     & 0.47 & 0.58 & 0.00 & 0.08 & \underline{0.70} & \underline{0.70} & 0.68 & 0.67 & 0.48 \\
Llama2-13B    & 0.66 & 0.45 & 0.02 & 0.09 & \underline{0.70} & \underline{0.70} & 0.70 & \underline{0.71} & 0.50 \\
OLMO-7B       & 0.57 & 0.56 & 0.02 & 0.15 & 0.68 & 0.69 & 0.68 & 0.68 & 0.50 \\
Gemma-7B      & 0.02 & 0.02 & 0.00 & 0.00 & 0.05 & 0.05 & 0.02 & 0.01 & 0.02 \\
BLOOM-7B      & 0.66 & 0.66 & 0.02 & \bf 0.30 & 0.68 & 0.67 & \bf 0.72 & \bf 0.72 & 0.55 \\ \midrule
CommandR-105B & 0.62 & 0.40 & \bf 0.04 & 0.09 & \underline{0.70} & \underline{0.70} & 0.67 & 0.67 & 0.49 \\
Llama2-70B    & 0.23 & 0.15 & 0.00 & 0.15 & \underline{0.70} & \underline{0.70} & \bf 0.72 & \underline{0.71} & 0.42 \\
Llama3-70B    & 0.55 & 0.34 & 0.02 & 0.13 & \bf 0.71 & \bf 0.71 & 0.66 & 0.66 & 0.47 \\ \midrule
GPT3.5-Turbo  & 0.57 & 0.34 & 0.02 & 0.09 & \bf 0.71 & \bf 0.71 & 0.66 & 0.66 & 0.47 \\
GPT-4o         & 0.62 & 0.79 & \underline{0.03} & 0.08 & \bf 0.71 & \bf 0.71 & 0.65 & 0.65 & 0.53 \\ 
%\midrule
\bottomrule
\end{tabularx}
\caption{Overview of instruction-tuned LLM performances evaluated through \textsc{SpeciaLex} for capturing the \textbf{\textsc{C3 (Target Audience)}} constraint where test instances were derived from the \textbf{Oxford 5000 lexicon} for CEFR. Each section division corresponds to the grouped LLMs based on similar scales. Values in bold mean the highest performance, while those underlined are second. Column $\mu$ denotes the mean performance across all subtasks. The underlined value for \textbf{Llama3-8B} denotes that it is the overall best-performing model for tasks requiring generated content aligned with the specified constraint. Column names can be referenced through subtask IDs in Table~\ref{tab:SpeciaLex-full-table}.}
\label{tab:SpeciaLex-result-cefr}
\end{table*}

\section{Experiments with \textsc{SpeciaLex}}
\label{sec:SpeciaLex-experiments}

\subsection{Models}
For \textsc{SpeciaLex}, we evaluated a diverse family of publicly accessible instruction-tuned models available on Huggingface. For models within the range of 1B-2B, we explored Gemma \cite{team2024gemma}, OLMO \cite{groeneveld2024olmo}, and BLOOM \cite{lescao2023bloom}. For models within the 7B to 13B, we included the Llama family \cite{touvron2023llama,touvron2023llama2}, Mistral \cite{jiang2023mistral}, as well as the larger versions OLMO and Gemma. For even larger models, we explored the 70B of Llama2 and Llama3 as well as Cohere's Command R with 105B. For commercial models, we explored GPT-3.5-Turbo and GPT-4o. Additional information on setup and hyperparameter can be found in Appendix~\ref{app:model_hyperparameter}.

\subsection{Performances on \textsc{SpeciaLex}'s Structured Prediction Tasks}
We highlight a number of insights by observing the performances of LLMs for structured prediction and classification from \textbf{\textsc{Checking}} and \textbf{\textsc{Identification}} tasks reported in Table~\ref{tab:SpeciaLex-result-ste} and Table~\ref{tab:SpeciaLex-result-cefr}. We refer the reader to Table~\ref{tab:SpeciaLex-full-table} in the Appendix~\ref{app:appendix} for the task number references throughout this section.

From the STE lexicon-based constraints, we see a straightforward trend in performance where the best models for capturing \textsc{C1} and \textsc{C2} are GPT-4o and GPT3.5-Turbo (ID1, ID2, and ID4). Llama3-70B has the closest runner-up performance for open models and obtains the best score for \textsc{Identification} with \textsc{C1} (ID3). On the other hand, for the target audience constraint \textsc{C3}, the best-performing models are open models, where the mid-sized Llama3-8B model obtains the three highest performance for \textsc{Checking} with full and minimal conformity and \textsc{Identification} which the latter ties with BLOOM-1B (ID11, ID12, and ID14). 

Through a paired $t$-test, we find no significance ($p > 0.05$, $t = 0.794$) in the performance difference of Llama3-70B against GPT-4o and GPT3.5-Turbo for \textsc{Checking} and \textsc{Identification} tasks capturing \textsc{C1} and \textsc{C2} constraints. Meanwhile, we do find significance with Llama3-8B against GPT-4o and GPT3.5-Turbo for target audience constraint \textsc{C3} ($p < 0.05$, $t = 0.015$) in favor of Llama3-8B a higher mean value ($0.55$ > $0.31$). These findings suggest that \textbf{open models like Llama3 can serve as strong, viable alternatives for content generation with structured lexicon-based constraints} if commercial models are unavailable or not within funding capacity.

\subsection{Performances on \textsc{SpeciaLex}'s Open-Ended Generation Tasks}
We highlight a number of insights by observing the performances of LLMs for open-ended generation from \textbf{\textsc{Rewriting}} and \textbf{\textsc{Open Generation}} tasks as reported in Table~\ref{tab:SpeciaLex-result-cefr} and Table~\ref{tab:SpeciaLex-result-cefr}.

Similar to the structured prediction tasks of \textsc{Checking} and \textsc{Identification}, we see favorable performances of commercial models GPT-4o and GPT-3.5-Turbo taking the top spots for \textsc{Open Generation} and \textsc{Rewriting}, particularly with on \textsc{C1} and \textsc{C2} constraints (ID8, ID9, and ID10) and on \textsc{C1} and \textsc{C3} with full and minimal conformity (ID15 and ID16). For open models, we see multiple models obtaining tied high performances. This includes Llama2-70B and BLOOM-7B together for \textsc{Open Generation} with full conformity (ID17), Llama3-70B and GPT-4o for \textsc{Open Generation} on \textsc{C1} (ID8) and on \textsc{Rewriting} with \textsc{C3} on full and minimal conformity (ID15 and ID16). 

%For the paired $t$-test, we aggregated performances of top open models BLOOM-7B, Llama2-70B, and Llama3-70B against the commercial models GPT-3.5-Turbo and GPT-4o. 
For the \textsc{Rewriting} and \textsc{Open Generation} tasks using STE lexicon-based constraints, we obtain no significance in performances of open models vs. commercial models ($p > 0.05$, $t = 0.150$). On the other hand, for \textsc{Rewriting} and \textsc{Open Generation} tasks using target audience constraints, we arrive at a significance ($p < 0.05$, $t = 0.021$) in favor of open models such as Llama3-70B with higher mean value ($0.70$ > $0.68$). With this, we further strengthen our previous findings and conclude that \textbf{open models like Llama2-70B, Llama3-70B, and BLOOM-7B remain competitive for controlled open-ended generation tasks} as first-choice models regardless of access to closed commercial models.

%\subsection{Current Model Standings}
%As shown in Tables~\ref{tab:SpeciaLex-result-ste} and \ref{tab:SpeciaLex-result-cefr}, the top performing LLMs for the \textsc{Specific Roles} and \textsc{Special Definitions} constraints derived from the STE lexicon are GPT-4o with an average of $0.82$ across the board followed by GPT-3.5-Turbo with $0.78$ and Llama3-70B with $0.76$. For the \textsc{Target Audience} constraint from the Oxford 5000 Wordlist, Llama3-8B places first with an average of $0.62$ followed by BLOOM-1B with $0.59$ and BLOOM-7B with $0.55$. 

\subsection{Error Analysis on Low-Performance Tasks}
We take a closer look at the tasks with generally poor performances from models. This is particularly evident for tasks in Table~\ref{tab:SpeciaLex-result-cefr} specifically on both \textsc{Idenfitication} subtasks requiring listing words from a text that are not recognized within the target audience level (ID13) and identifying the correct target audience level (ID14). For the former, upon manual error analysis of model outputs, \textbf{LLMs evaluated for the subtask often provide an insufficient number of required words} (e.g., only giving $1-3$ words while the required is $5-6$), which includes words that are already within the recognized target audience level. For the latter, we see a trend where \textbf{LLMs tend to oversimplify their estimations to lower levels} (e.g., the correct level is B2, but models will give A2 or A1). We find similar insights from previous works on instruction-tuned LLMs oversimplifying level estimations for in-context learning tasks \cite{imperial-tayyar-madabushi-2023-flesch}. 

In hindsight, knowing that commercial and open LLMs may underperform for niche tasks that specific NLP tools or models can easily solve is something that domain-specific users should know when using these LLMs. Thus, we see this as an advantage that our benchmark exposes certain limitations that can inform the NLP community to build upon this work. We reserve the improvement of LLM performance for these specific subtasks for future work.

%%
%% SCALE EXPERIMENT
%%

\begin{figure}[!t]
    \centering
    \begin{subfigure}[t]{0.23\textwidth}
        \includegraphics[width=\textwidth,trim={0.5cm 0.5cm 0.5cm 0.5cm}, clip]{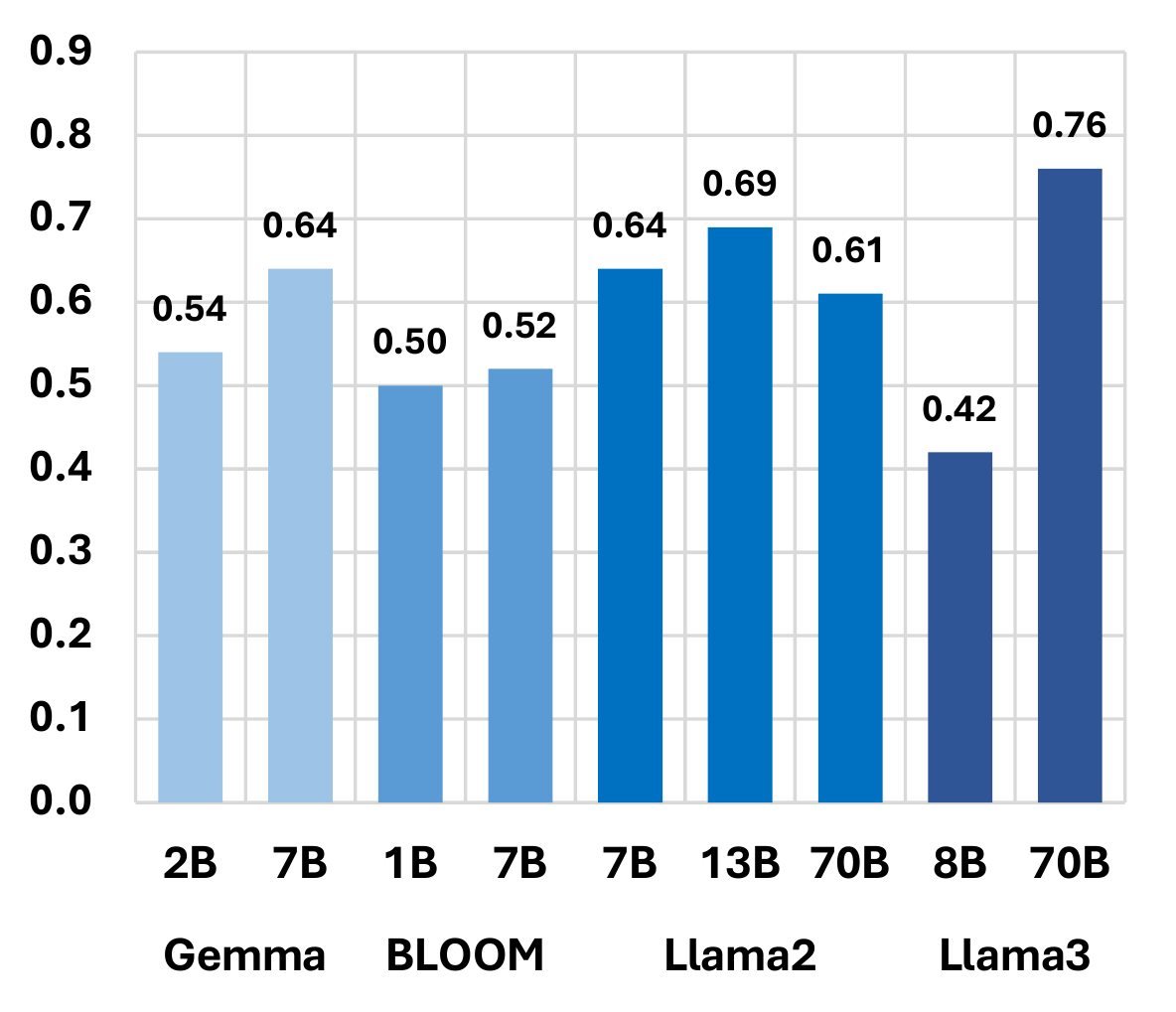}
    \end{subfigure}
    \hspace{0.0cm}
    \begin{subfigure}[t]{0.23\textwidth}
        \includegraphics[width=\textwidth,trim={0.5cm 0.5cm 0.5cm 0.5cm}, clip]{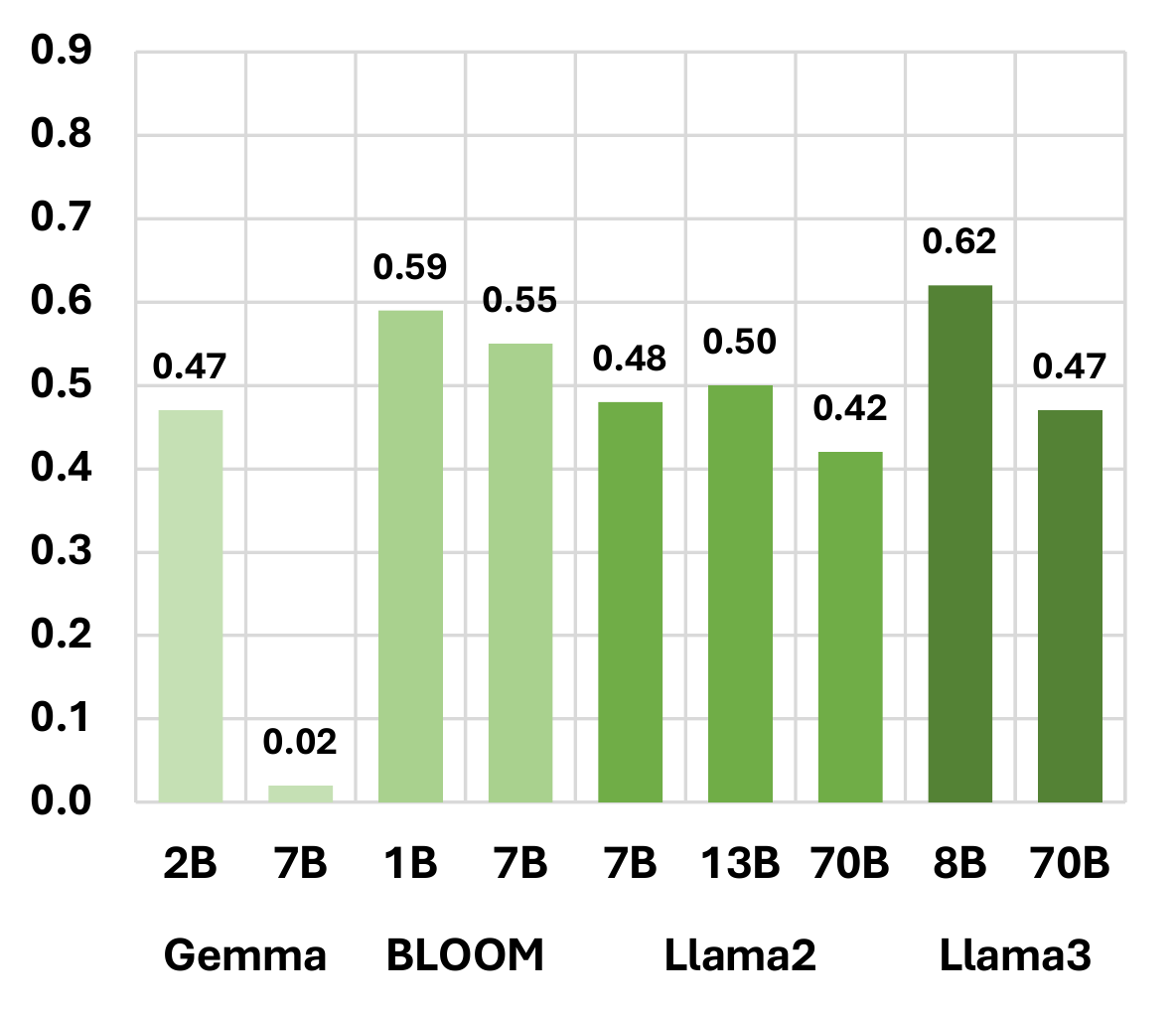}
    \end{subfigure}
    \vspace{0pt}
    \caption{Mean model performances based on increasing model scale. We report performances of models for STE-based lexicon constraints (\textbf{left}) as seen in Table~\ref{tab:SpeciaLex-result-ste} while the Oxford 5000 lexicon for CEFR-based constraints (\textbf{right}) as seen in Table~\ref{tab:SpeciaLex-result-cefr}. We observe an obvious growth trend in STE performance for larger models while a notable advantage in smaller models for the CEFR.}
    \label{fig:model-scale}
\end{figure}

\section{A \textsc{SpeciaLex} Guide}
In this section, we outline a number of important points for consideration to guide researchers in using \textsc{SpeciaLex} as a reference or an evaluation tool for specific domain data and constraints.\\

\noindent \textbf{Do bigger models have better performance? \ul{It depends on the task.}} It is a common observation from empirical experiments with LLMs that the larger the scale, the higher the generalization and performance across diverse tasks \cite{weiemergent2022,wei2021finetuned,brown2020language}. However, the choice of larger models may be expensive and impractical for domain adaptation, where performance on a limited set of tasks (or even a singular task) is often prioritized. Upon aggregating the mean results from Tables~\ref{tab:SpeciaLex-result-ste} and ~\ref{tab:SpeciaLex-result-cefr} of models with increasing scale in Figure~\ref{fig:model-scale}, we see only favorable performance for larger models on STE-based constraints focused on specific POS and special definitions. In the case of using target audience constraint, we observe that even the 8B version of Llama3 is better than all other models tested. Thus, we recommend researchers consider the nature of the task first, as smaller models have empirically shown to be able to achieve comparable performance on select constraints. \\

\noindent \textbf{Are open models good enough? \ul{Yes.}} While it is also a common notion that commercial models such as GPT-4 by OpenAI are popularly known and advertised as the go-to standard for general NLP tasks, we provide empirical evidence in this study that open models are equally as performant and can serve as a practical alternative for the research community. Revisiting our findings from Section~\ref{sec:SpeciaLex-experiments}, open models such as Llama3-8B and 70B are able to achieve comparable---if not higher in some cases--performances across the four core tasks based on mean scores. \\

\noindent \textbf{Do high-quality training data and model recency matter? \ul{Yes.}} Model scale may not be the only signal of effectiveness for capturing lexicon-based constraints. We recommend weighing the quality of data used for training the LLMs and using the most recent model versions released by their research developers. We see this particular advantage in the Llama family models with $15$T token count used for pre-training data as well as using high-quality data filters\footnote{\url{https://ai.meta.com/blog/meta-llama-3/}} powered by Llama2. With this advantage, Llama3 was able to achieve generally higher task performances in \textsc{Specialex} than Llama2. Likewise, we posit that Llama3's recency among all the other models may have given certain advantages in terms of data quality through scoping more and larger published open-source datasets used for pre-training. \\

\begin{figure}[!t]
    \centering
    \begin{subfigure}[t]{0.23\textwidth}
        \includegraphics[width=\textwidth,trim={0.5cm 0.5cm 0.5cm 0.5cm}, clip]{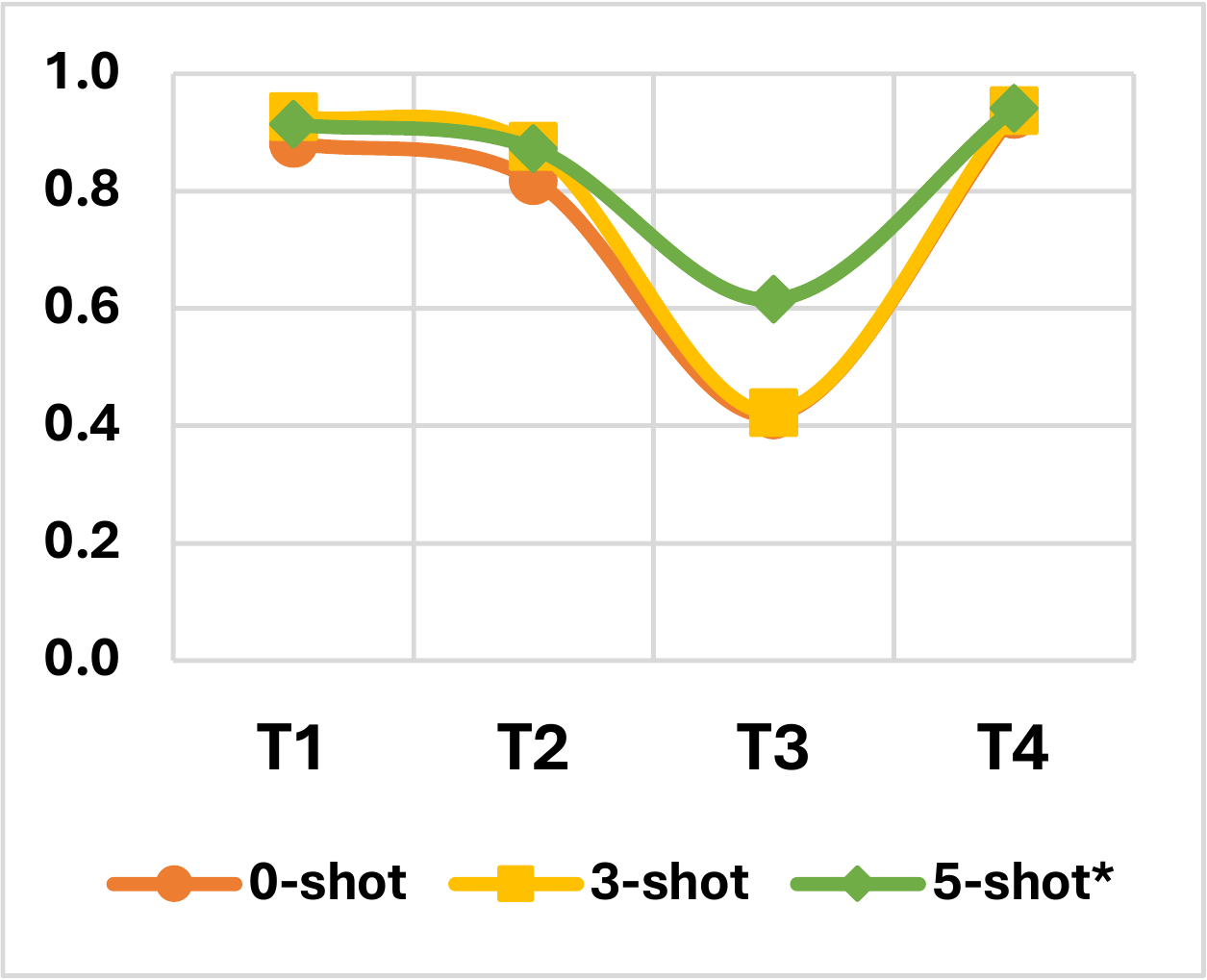}
    \end{subfigure}
    \hspace{0.0cm}
    \begin{subfigure}[t]{0.23\textwidth}
        \includegraphics[width=\textwidth,trim={0.5cm 0.5cm 0.5cm 0.5cm}, clip]{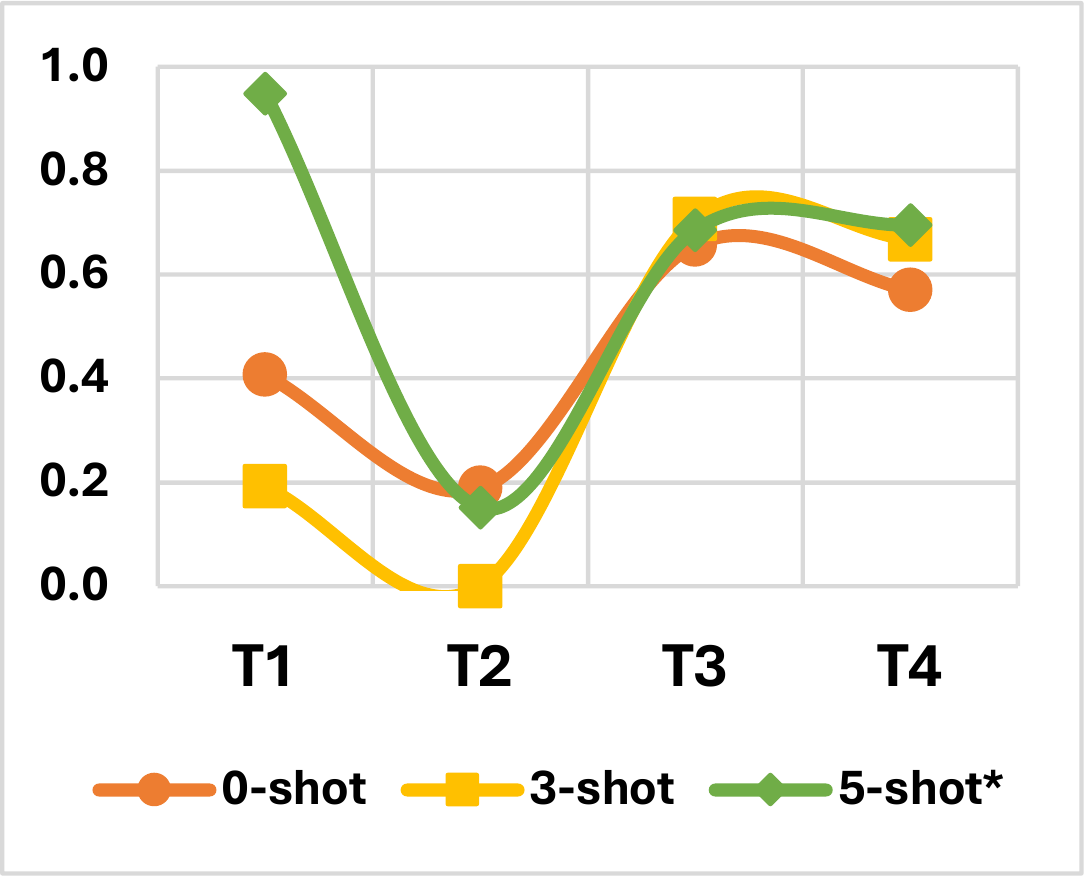}
    \end{subfigure}
    \vspace{0pt}
    \caption{Mean performances of based on various few-shot ICL demonstrations per task category. We use the best-performing models from the STE and Oxford 5000 lexicon constraints, which are GPT-4o (\textbf{left}) and Llama3-8B (\textbf{right}), respectively. We observe generally higher performance using the standard 5-shot approach on all the core tasks, denoting the effectivity of providing higher quality examples for ICL.}
    \label{fig:model-fewshot}
\end{figure}

\noindent \textbf{How many demonstrations do I need for ICL? \ul{Five is a good start.}} \textsc{SpeciaLex} benchmarks models via in-context learning since prompting and providing additional information and target output is the most common way of interacting and delegating tasks to LLMs. As such, we recommend starting with around five or more diverse demonstrations rather than a zero-shot method for lexicon-based constraints to maximize the effectiveness of in-context learning. We support this recommendation by exploring various few-shot techniques from the best-performing models for STE and CEFR-based constraints, as seen in Figure~\ref{fig:model-fewshot}. From the experiment, we report that using the standard 5-shot setup done in the major experiments in Table~\ref{tab:SpeciaLex-result-ste}
 and \ref{tab:SpeciaLex-result-cefr} generally obtain better performance than its equivalent lower shot examples.\\

\noindent \textbf{Is in-context learning better than domain-specific finetuning? \ul{ICL allows flexibility and preserves general model performance}.} The benefit of \textsc{SpeciaLex} by benchmarking via in-context learning is that it avoids re-training LLMs to one specific task only and preserves the user’s ability to use the LLM (or LLM interfaces such as ChatGPT) to perform other downstream tasks such question answering, summarizing content, and solving problems to name a few. Moreover, in case LLMs perform poorly on tasks via in-context learning, the results would serve as a helpful direction for domain users to then explore finetuning or other optimization methods. Nonetheless, this recommendation is not prescriptive if domain users' priority is to develop a model that performs well in capturing constraints from only one source of reference via the specialized lexicon.

%Our motivation for developing SpeciaLex draws from the same motivation as any LLM benchmark: to further the understanding of the limitations of LLMs, particularly for domain-specific constraints. Knowing that commercial and open LLMs may underperform for niche simple tasks that specific NLP tools can solve is something that users in the domain should know when using these LLMs. Thus, we see this as an advantage that our benchmark exposes certain limitations that can inform the NLP community to build upon this work. We will include a more in-depth discussion in the SpeciaLex Guide section on how to improve low-performance tasks through better domain-aligned training steps for future work.

%%
%% FEW SHOT EXPERIMENT
%%

\section{Conclusion}
In this work, we introduced \textsc{SpeciaLex}, a benchmark for evaluating state-of-the-art LLMs in capturing specialized lexicon-based constraints for content generation tasks commonly prevalent across interdisciplinary areas such as education, technical writing, and engineering. We provided an in-depth and empirical exploration of model performance, including looking at the effects of model scale, openness, few-shot setup, and recency. Our findings support the use of open models such as Llama8-3B as good, competitive starting resources for the benchmark. In hindsight, \textsc{SpeciaLex} will serve as a reference guide for researchers within and outside of the NLP community, where they can check what specific models are good for certain task types (checking, identification, rewriting, generation) or handle specific constraints (specific roles, special definitions, target audience) and springboard further research based on their own domain specifications.

%which also serves as a step forward for accessible community adoption and springboarding to various domains. 

%In hindsight, \textsc{SpeciaLex} will serve as a reference guide for researchers within and outside of the NLP community, where they can check what specific models are good for certain tasks (checking, identification, rewriting, generation) or handle specific constraints (specific roles, special definitions, target audience) and springboard further research based on their own domain specifications.

\section*{Limitations}

\noindent \textbf{Application to Multilingual Domain}. Our work, including the data resources we used for building \textsc{SpeciaLex} tasks and the LLMs we evaluated, mainly focuses on the English language. We do not claim that the performances of the models we reported in this paper will be comparable to tasks where the source of lexicon-based constraints is in a different language. Investigating the capabilities of LLMs in capturing multilingual lexicon-based constraints is a research opportunity left for future work.\\

\noindent \textbf{Coverage of Non Lexicon-Based Constraints}. For uniformity of experiment setups and achieving a centralized benchmark, our work specifically focuses on evaluating to what extent LLMs can capture lexicon-based constraints via in-context learning. Thus, we do not focus on evaluating rules beyond those covered by a specialized lexicon. For example, in Simple Technical English (STE), although not part of the lexicon, there are some additional recommended rules on phrasing, such as \textit{maintaining only one topic per paragraph} or \textit{start an instruction with a descriptive statement (dependent phrase or clause)}. Upon recommendation by the experts we collaborated with, we did not include these rules in the experiment process.\\

\noindent \textbf{Evaluation via In-Context Learning}. In this work, we used prompting through in-context learning as one of the easiest ways users of various domain areas use an LLM (or LLM interfaces such as ChatGPT) with minimal effort. Moreover, the benefit of benchmarking via in-context learning is that it avoids fine-tuning or re-training the LLM to perform one specific task only and still preserves the LLM’s ability to perform language tasks such as summarizing, chatting, and answering questions. Lastly, evaluating through in-context learning can be considered the first step towards exposing the limitations of LLMs (as done by previous benchmarks) which can serve as a springboard for further domain-specific training or finetuning in future works. \\

\section*{Ethics Statement}
This work used LLMs for the generation of texts to conform to lexicon-based constraints derived from Simple Technical English and Oxford 5000, which are existing publicly accessible expert-developed corpora provided proper acknowledgments. The prompts crafted for each subtask of the \textsc{SpeciaLex} benchmark are all derived from the two mentioned data sources and do not instruct the LLMs to explicitly nor implicitly produce harmful texts. Overall, we do not see any serious ethical implications from this work.

\section*{Acknowledgements}
We would like to thank Brian North and Orlando Chiarello for the insightful discussions on capturing CEFR and ASD-STE standards used in this work. ASD-STE100 Simplified Technical English is a Copyright and a Trademark of ASD, Brussels, Belgium. This work made use of the Hex GPU cloud of the Department of Computer Science at the University of Bath. JMI is supported by the National University Philippines and the UKRI Centre for Doctoral Training in Accountable, Responsible and Transparent AI [EP/S023437/1] of the University of Bath.

% Bibliography entries for the entire Anthology, followed by custom entries
%\bibliography{anthology,custom}
% Custom bibliography entries only
\bibliography{references,anthology}

\clearpage
\appendix

\section{Appendix}
\label{app:appendix}
In the following sections, we provide additional information, such as examples and statistics regarding the datasets, experiment procedures, and tasks used for building the \textsc{SpeciaLex} benchmark.

\section{Model Hyperparamter and Generation Setting}
\label{app:model_hyperparameter}
Implementation-wise, we used Huggingface's Inference API ({\url{https://huggingface.co/inference-api/serverless}}) and Text Generation Pipeline for these models and set temperature to $0.0$ for all tasks in line with the deterministic nature and max tokens to $300$ for \textsc{Rewriting} and \textsc{Open Generation} tasks. For running the models for inference, we used our university's GPU cloud server with 8 NVIDIA GeForce RTX 3090 with 24GB memory size. For closed commercial models, we evaluated GPT3.5-Turbo and GPT-4o for comparison with a January 25 and May 2024 knowledge cutoff, respectively, using OpenAI's API ({\url{https://openai.com/api/}}). We omit OLMO-1B in Table ~\ref{tab:SpeciaLex-result-cefr} due to the generation of gibberish texts for this setup.

\section{Additional Information on Datasets}
\label{app:datasets}

\subsection{Oxford 5000}
We provide additional statistical information regarding the Oxford 5000 lexicon used for \textsc{SpeciaLex}. Table~\ref{tab:cefr_breakdown_percentage} shows the breakdown of the number of unique words associated per target audience level of the CEFR scale used for Oxford 5000. Since the nature of CEFR is ordinal in practice (e.g., a B1 learner recognizes words from previous levels such as A2 and A1), we combined the words per category successively when evaluating for density of content words in the custom lexicon-matching script done from the experiments in Table~\ref{tab:SpeciaLex-result-cefr}. We also provide an example of $25$ words that experts found to be recognizable per target audience level in Table~\ref{tab:cefr_sample_words}.

\begin{table}[!htbp]
\centering
\begin{tabular}{@{}lrrrrr@{}}
\toprule
      & \bf A1     & \bf A2     & \bf B1     & \bf B2     & \bf C1     \\ \midrule
Count & 897    & 867    & 838    & 1,422  & 1,311  \\
\%    & 16.8 & 16.5 & 15.7 & 26.6 & 24.5 \\ \bottomrule
\end{tabular}
\caption{Breakdown of number of words and percentage of each CEFR level from the Oxford 5000 lexicon.}
\label{tab:cefr_breakdown_percentage}
\end{table}

\subsection{STE}
We provide additional statistical information regarding the Simple Technical English (STE) lexicon used for \textsc{SpeciaLex}. Table~\ref{tab:ste_breakdown_percentage} shows the breakdown of original words, it's a corresponding recommended alternative with correct role or POS information, and words with special definitions per POS category recognized by the lexicon. As such, the data from the first two columns were used for building the tasks for \textsc{C1 - Specific Role} and the third for \textsc{C2 - Special Definition}. For this study, we used the 2017 version provided by the STEMG representatives we collaborated with, which is the previous version to the current 2021 version available to download from the official website ({\url{https://www.asd-ste100.org/}}). This is due to embargo restrictions on machine-readable copies. Furthermore, we obtained explicit permission from the STEMG representatives to share the transformed version of the STE lexicon with respect to benchmark tasks to be shared as a research artifact of this work.

\begin{table}[!htbp]
\centering
\begin{tabular}{@{}lrr|r@{}}
\toprule
\bf POS   & \bf Original & \bf Alternative & \bf Special Def \\ \midrule
NOUN  & 212      & 276         & 243                \\
VERB  & 648      & 590         & 235                \\
ADP   & 27       & 39          & 49                 \\
ADJ   & 269      & 247         & 254                \\
ADV   & 85       & 80          & 118                \\
SCONJ & 12       & 22          & 18                 \\
PRON  & 6        & 4           & 19        \\ \bottomrule        
\end{tabular}
\caption{Breakdown of original words, corresponding alternatives, and words with special definitions per POS category from the STE lexicon.}
\label{tab:ste_breakdown_percentage}
\end{table}

\section{Additional Information on Constructing Prompts for Tasks}
\label{app:additional_task_infor}
For tasks covering \textsc{C1 - Specific Role} and \textsc{C2 - Special Definition}, the information required to build the prompts for their associated tasks was all derived from what is available in the STE lexicon as seen in Table~\ref{tab:ste_approved_pos_examples} and Table~\ref{tab:ste_approved_definition_examples}. For example, for Task ID5, we want to prompt an LLM to rewrite a sentence so that the target word is replaced by its STE-approved alternative and POS information. Thus, we only need to get data from the incorrect sentence column, the target word column and its POS, and the approved word column and its POS to build the prompt, which we can see in Figure~\ref{fig:task6}.

For tasks covering \textsc{C3 - Target Audience}, unlike STE, Oxford 5000, the lexicon does not come with pre-compiled examples of stories conforming to each specific target audience level. Thus, we use an external data source for this, which is the \textsc{TinyStories} corpus \cite{eldan2023tinystories}, which is a GPT-4 generated compilation of short stories. The selection of this corpus is due to its recency and obtaining high qualitative evaluation in terms of consistency, grammar, creativity, and plot by human annotators  \cite{eldan2023tinystories}. Using our custom lexicon-matching script, we select entries from the \textsc{TinyStories} corpus that fit each target audience category in the CEFR levels recognized by the Oxford 5000 lexicon and use them according to task requirements. For example, in Task ID14 in Figure~\ref{fig:task14}, we used \textsc{TinyStories} entries classified under different CEFR levels to prompt an LLM to guess their correct CEFR level, given a few examples for in-context learning. Another example in Task ID15 in Figure~\ref{fig:task15}, we prompt an LLM to rewrite the story to a target lower or higher audience level.

\section{Additional Information on Evaluation Methods}
\label{app:evaluation}
We provide additional information about the evaluation methods used for the constraints. For the \textsc{C2 - Special Definition} constraint where GPT-4 is used as the judge, we use the following prompt template below: 

\begin{figure}[!htbp]
    \begin{tcolorbox}[colframe=yellow, colback=white, coltitle=white, center title, fonttitle=\bfseries]
    \small
    
    Sentence: \{\{sentence\}\}\\
    Word: \{\{word\}\}\\
    Approved Definition: \{\{approved\_definition\}\}\\

    Given the information above, judge if the given word is used in the sentence with respect to its approved definition. Answer directly with YES or NO. 
    
    \end{tcolorbox}
    \caption{Prompt template for using GPT-4 as a judge to evaluation the \textsc{Special Definition (C2)} constraint.}
    \label{fig:gpt-as-judge}
\end{figure}

For the \textsc{C3 - Target Audience} constraint, the formula used for the lexicon-matching script is as follows:

\begin{equation}
\text{score} = \frac{\sum_{w \in t} \mathbbm{1}({w \in L_i})}{n}
\end{equation}

where $w$ denotes each content word from the text $t$ being evaluated for occurrence in the set of words recognized by the target audience level $L_i$ (e.g., A2) and normalized by the total number of words $n$ of the text. $\mathbbm{1}$ is an indicator function that counts $1$ for each match. As mentioned, closer values to $1.0$ are better since they denote texts with a higher density of words recognized by the specific target audience level.

\section{Task Prompt Templates}
\label{app:task_prompt_templates}
We provide the base prompt templates used for each task from \textsc{SpeciaLex} in the last portion of this document from Figures~\ref{fig:task1} to \ref{fig:task18}. The templates were adopted from previous benchmark tasks such as \textsc{LegalBench} \cite{guha2024legalbench} and \textsc{RAFT} \cite{alex2021raft} where few-shot examples are also used for in-context learning. The template visualizations are color-coded with respect to the task: \colorbox{teal}{Teal} for \textsc{Checking}, \colorbox{purple}{Purple} for \textsc{Idenfitication}, \colorbox{violet}{Violet} for \textsc{Rewriting}, and \colorbox{cyan}{Cyan} for \textsc{Open Generation}.

%%
%% TABLE - FULL TASK DESCRIPTION
%%

\setlength{\tabcolsep}{10pt}

\begin{table*}[!t]
\small
\centering
\renewcommand{\arraystretch}{1.3}
\begin{tabular}{@{}p{0.5cm}p{5.5cm}p{0.5cm}p{1cm}p{1cm}p{1.7cm}p{1.5cm}@{}}
\toprule
\bf ID&
  \bf Task Description&
  \bf Task&
  \bf Constraint&
  \bf Corpora&
  \bf Evaluation&
  \bf Instances\\ \midrule
1 &
  Given a word and a text, check if the word is used according to its approved POS. &
  \sc T1 &
  \sc C1 &
  STE &
  Exact Acc &
  72 \\
2 &
  Given a word and a text, check if the word is used according to its approved definition. &
  \sc T1 &
  \sc C2 &
  STE &
  Exact Acc &
  64 \\
3 &
  Given a text, identify the word that is incorrectly used according to approved definition. &
  \sc T2 &
  \sc C1 &
  STE &
  Exact Acc &
  69 \\
4 &
  Given a text, identify the word that is incorrectly used according to approved POS. &
  \sc T2 &
  \sc C2 &
  STE &
  Exact Acc &
  77\\
5 &
  Given a text and a word, rewrite the text so that the word is replaced by its approved substitute and POS. &
  \sc T3 &
  \sc C1 &
  STE &
  POS Evaluator &
  300 \\
6 &
  Given a text and a word, rewrite the text so that the word is replaced by its approved substitute and definition. &
  \sc T3 &
  \sc C2 &
  STE &
  GPT-4 &
  82 \\
7 &
  Given a text and a word, rewrite the text so that the word is used according to its approved substitute, definition, and POS. &
  \sc T3 &
  \sc C1, C2 &
  STE &
  POS Evaluator, GPT-4 &
  67 \\
8 &
  Given a word, generate a text where the word is used according to its approved POS. &
  \sc T4 &
  \sc C1 &
  STE &
  POS Evaluator &
  175\\
9 &
  Given a word, generate a text where the word is used according to its approved definition. &
  \sc T4 &
  \sc C2 &
  STE &
  GPT-4 &
  175\\
10 &
  Given a word, generate a text where the word is used according to its approved definition and POS. &
  \sc T4 &
  \sc C1, C2 &
  STE &
  POS Evaluator, GPT-4 &
  175 \\
11 &
  Given a text and a target audience via a category, check if all words in the text that occur within the category. &
  \sc T1 &
  \sc C3 &
  Oxford 5000 &
  Exact Acc &
  53 \\
12 &
  Given a text and a target audience via a category, check if 95\% of content words in the text occur within the category. &
  \sc T1 &
  \sc C3 &
  Oxford 5000&
  Exact Acc &
  62\\
13 &
  Given a text and target audience via a category, identify all words in the text that occur beyond the category. &
  \sc T2 &
  \sc C3 &
  Oxford 5000&
  Exact Acc &
  55 \\
14 &
  Given a text, identify the correct target audience via selecting a category. &
  \sc T2 &
  \sc C3 &
  Oxford 5000&
  Exact Acc &
  53 \\
15 &
  Given a text and a target audience via a category, rewrite the text where all of its content words belong to the category. &
  \sc T3 &
  \sc C3 &
  Oxford 5000&
  Dictionary Match &
  53 \\
16 &
  Given a text and a target audience via a category, rewrite the text where at least 95\% of its content words belong to the category. &
  \sc T3 &
  \sc C3 &
  Oxford 5000&
  Dictionary Match &
  53\\
17 &
  Given a topic prompt and a target audience via a category, generate a text where all of its content words belong to the category &
  \sc T4 &
  \sc C3 &
  Oxford 5000&
  Dictionary Match &
  100\\
18 &
  Given a topic prompt and a target audience via a category, generate a text where at least 95\% of its content words belong to the target. &
  \sc T4 &
  \sc C3 &
  Oxford 5000&
  Dictionary Match &
  100\\ \bottomrule
\end{tabular}
\caption{Full details of the $18$ tasks covered by \textsc{SpeciaLex} distributed across $4$ core tasks (\textsc{Checking}, \textsc{Identification}, \textsc{Rewriting}, and \textsc{Open Generation}) and $3$ lexicon-based constraints (\textsc{Specific Role}, \textsc{Special Definition}, \textsc{Target Audience}) from Simple Technical English (STE) and Oxford 5000 for CEFR. The number of test instances total to $1,785$.}
\label{tab:SpeciaLex-full-table}
\end{table*}

%%
%% TABLE - OXFORD 5000 WORDLIST EXAMPLES
%%

\begin{table*}[!t]
\centering
\begin{tabular}{@{}ccccc@{}}
\toprule
\bf A1              & \bf A2                     & \bf B1                     & \bf B2                     & \bf C1                      \\ \midrule
\textit{above}  & \textit{asleep}        & \textit{absolutely}    & \textit{accurate}      & \textit{abolish}        \\
\textit{across} & \textit{appear}        & \textit{academic}      & \textit{acknowledge}   & \textit{accumulation}   \\
\textit{ask}    & \textit{average}       & \textit{achievement}   & \textit{acquire}       & \textit{activist}       \\
\textit{big}    & \textit{behavior}      & \textit{battery}       & \textit{blind}         & \textit{battlefield}    \\
\textit{bike}   & \textit{blood}         & \textit{border}        & \textit{broadcast}     & \textit{biography}      \\
\textit{cake}   & \textit{celebrity}     & \textit{careless}      & \textit{bacteria}      & \textit{bureaucracy}    \\
\textit{call}   & \textit{coast}         & \textit{concentrate}   & \textit{commission}    & \textit{classification} \\
\textit{cold}   & \textit{complain}      & \textit{countryside}   & \textit{complicated}   & \textit{collaboration}  \\
\textit{dark}   & \textit{designer}      & \textit{documentary}   & \textit{contemporary}  & \textit{configuration}  \\
\textit{day}    & \textit{disaster}      & \textit{disadvantaged} & \textit{deeply}        & \textit{destructive}    \\
\textit{dear}   & \textit{disease}       & \textit{discount}      & \textit{deliberate}    & \textit{detection}      \\
\textit{egg}    & \textit{engineer}      & \textit{environmental} & \textit{dishonest}     & \textit{deteriorate}    \\
\textit{eat}    & \textit{experience}    & \textit{exchane}       & \textit{emphasize}     & \textit{electoral}      \\
\textit{ear}    & \textit{experiment}    & \textit{frightened}    & \textit{examination}   & \textit{empirical}      \\
\textit{face}   & \textit{fortunately}   & \textit{friendship}    & \textit{fundamental}   & \textit{favorable}      \\
\textit{fast}   & \textit{furniture}     & \textit{headache}      & \textit{facility}      & \textit{forthcoming}    \\
\textit{fish}   & \textit{foreign}       & \textit{hockey}        & \textit{landscape}     & \textit{ideological}    \\
\textit{fire}   & \textit{fiction}       & \textit{lorry}         & \textit{logical}       & \textit{ironically}     \\
\textit{girl}   & \textit{government}    & \textit{loudly}        & \textit{military}      & \textit{legislative}    \\
\textit{hair}   & \textit{hero}          & \textit{lifestyle}     & \textit{minister}      & \textit{literacy}       \\
\textit{half}   & \textit{habit}         & \textit{possibility}   & \textit{mysterio}      & \textit{mainstream}     \\
\textit{high}   & \textit{international} & \textit{poster}        & \textit{nevertheless}  & \textit{mobilize}       \\
\textit{juice}  & \textit{invention}     & \textit{profile}       & \textit{nightmare}     & \textit{niche}          \\
\textit{learn}  & \textit{mathematics}   & \textit{reception}     & \textit{occassionally} & \textit{newsletter}     \\
\textit{laugh}  & \textit{manager}       & \textit{relationship}  & \textit{obligation}    & \textit{nonsense}       \\ \bottomrule
\end{tabular}
\caption{Sample 25 unique words from the Oxford 5000 lexicon for each target audience category.}
\label{tab:cefr_sample_words}
\end{table*}

%%
%% TABLE - STE SPECIFIC ROLE EXAMPLES
%%
\setlength{\tabcolsep}{5pt}

\begin{table*}[!t]
\small
\centering
\renewcommand{\arraystretch}{1.3}
\begin{tabularx}{\textwidth}{@{}p{1.7cm}p{1.0cm}p{1.7cm}p{1.3cm}p{4.3cm}p{4.3cm}@{}}
\toprule
\bf Word        & \bf POS  & \bf Alternative & \bf POS & \bf Approved Example                                      & \bf Incorrect Example                                     \\ \midrule
\textit{abandon} &
  VERB &
  \textit{stop} &
  VERB &
  Stop the engine start procedure. &
  Abandon engine start. \\
\textit{abate} &
  VERB &
  \textit{decrease} &
  VERB &
  When the wind speed decreases to less than 30 knots, you can open the cargo door. &
  When the wind abates to less than 30 knots, you can open the cargo door. \\
\textit{abnormality} &
  NOUN &
  \textit{defect} &
  NOUN &
  Examine the seal for defects. &
  Examine the seal for abnormalities. \\
\textit{bank} &
  VERB &
  \textit{bank} &
  NOUN &
  The V-bars give the indication for a bank. &
  V-Bars indicate command to bank. \\
\textit{bolt} &
  VERB &
  \textit{bolt} &
  NOUN &
  Attach the track to the channels with the bolts. &
  Bolt track to channels. \\
\textit{break} &
  NOUN &
  \textit{stop} &
  VERB &
  If the transmission stops, cancel the test. &
  If there is a break in transmission, cancel the test. \\
\textit{calculation} &
  NOUN &
  \textit{calculate} &
  VERB &
  In this example, we only calculated the data applicable to a type B unit. &
  The data used for the calculations in this example apply only to a Type B unit. \\
\textit{care} &
  NOUN &
  \textit{precaution} &
  NOUN &
  Obey the safety precautions when you do work with high voltages. &
  You must take care when you work with high voltages. \\
\textit{centralize} &
  VERB &
  \textit{center} &
  NOUN &
  Set the controls to the center position. &
  Centralize the controls. \\
\textit{destroy} &
  VERB &
  \textit{unserviceable} &
  ADJ &
  Make the container unserviceable to make sure that you cannot use it again. &
  To avoid further use, destroy the container. \\
\textit{double} &
  ADJ &
  \textit{two} &
  NOUN &
  You must see two marks on the stand. &
  Double marks must appear on the stand. \\
\textit{earth} &
  VERB &
  \textit{ground} &
  VERB &
  Make sure that the fuel tanks are correctly grounded. &
  Make sure the fuel tanks are correctly earthed. \\
\textit{emit} &
  VERB &
  \textit{from} &
  ADP &
  The fumes from this material are dangerous to the skin. &
  The vapors that this material emits are dangerous to the skin. \\
\textit{factor} &
  NOUN &
  \textit{cause} &
  VERB &
  There can be many causes for corrosion. &
  Corrosion can be caused by several factors. \\
\textit{fatal} &
  ADJ &
  \textit{kill} &
  VERB &
  High voltage in the electronic system can kill you. &
  High voltage in the electronic system can be fatal. \\
\textit{finish} &
  VERB &
  \textit{complete} &
  VERB &
  Complete the test. &
  Finish the test. \\
\textit{gash} &
  VERB &
  \textit{damaged} &
  ADJ &
  If the thermal blanket is damaged, do repair no. 9. &
  If the thermal blanket is gashed, do repair No. 9. \\
\textit{gloss} &
  NOUN &
  \textit{shiny} &
  ADJ &
  Polish the surface until it is very shiny. &
  Polish the surface to a high gloss. \\
\textit{hold} &
  NOUN &
  \textit{hold} &
  VERB &
  Make sure that you hold the rod tightly. &
  Make sure that you have a tight hold on the rod. \\
\textit{impression} &
  NOUN &
  \textit{think} &
  VERB &
  If you think that a tire has low pressure, do the steps that follow: &
  If you have the impression that a tire has low pressure, do the steps that follow. \\
\textit{incline} &
  NOUN &
  \textit{slope} &
  NOUN &
  You can adjust the slope of the ramp. &
  You can adjust the incline of the ramp. \\
\textit{loop} &
  VERB &
  \textit{loop} &
  NOUN &
  Make a loop of wire around the unit. &
  Loop the wire around the unit. \\
\textit{lose} &
  VERB &
  \textit{decrease} &
  VERB &
  The effect of the solvent decreases quickly. &
  The solvent loses its effectiveness quickly. \\
\textit{mark} &
  VERB &
  \textit{identify} &
  VERB &
  Identify the component with a code to help you to install it again correctly. &
  Mark the component with a code that will facilitate its correct reinstallation. \\
\textit{medium} &
  ADJ &
  \textit{moderate} &
  ADJ &
  Apply moderate pressure. &
  A medium amount of pressure must be applied.\\ \bottomrule      
\end{tabularx}
\caption{Sample 25 entries from the STE lexicon containing words and their recommended alternatives with approved POS information, correct, and incorrect example sentences.}
\label{tab:ste_approved_pos_examples}
\end{table*}

%%
%% TABLE - STE SPECIAL DEFINITION EXAMPLES
%%

\begin{table*}[!t]
\centering
\small
\renewcommand{\arraystretch}{1.3}
\begin{tabularx}{\textwidth}{@{}p{2cm}p{1.5cm}p{6cm}p{5.5cm}@{}}
\toprule
\bf Word                    & \bf POS  & \bf Approved Definition                                             & \bf Approved Example                                                   \\ \midrule
\textit{abrasive}       & ADJ  & that can remove material by friction                            & Dust, when mixed with oil, has an abrasive effect.                \\
\textit{accept}         & VERB & to make a decision that something is satisfactory               & Accept the relay if it is serviceable.                            \\
\textit{aft}            & ADJ  & nearer to the rear of an air or sea vehicle                     & The pump is in the aft cell of the fuselage tank.                 \\
\textit{bend}           & NOUN & the area where something is bent                                & Examine the bends for cracks.                                     \\
\textit{bleed}          & VERB & to let a gas out of                                             & Bleed the speedbrake hydraulic system.                            \\
\textit{bond}           & VERB & to make an electrical bond                                      & The static discharger is electrically bonded to the frame.        \\
\textit{can} &
  VERB &
  helping verb that means to be possible, to be able to, or to be permitted to &
  A mixture of fuel and oxygen can cause an explosion. \\
\textit{control}        & NOUN & something that controls                                         & Use the manual control in an emergency.                           \\
\textit{device}         & NOUN & something used to do a task                                     & Install the safety devices.                                       \\
\textit{dim}            & ADJ  & not bright                                                      & During night operation, make sure that the panel lights are dim.  \\
\textit{divide}         & VERB & to separate into parts or groups                                & You can divide the drains into three primary groups.              \\
\textit{edge} &
  NOUN &
  a line that is the intersection of two surfaces of a solid object &
  The distance between the edge of the panel and the partition must not be more than 0.05 mm. \\
\textit{engage}         & VERB & to correctly align and come together                            & Engage the clutch.                                                \\
\textit{explosive} &
  ADJ &
  that can cause an explosion &
  The safety precautions that follow are applicable to explosive items. \\
\textit{finger-tighten} & VERB & tighten with your fingers                                       & Tighten the nut with your fingers.                                \\
\textit{flange}         & NOUN & an end surface at an angle                                      & Make sure that the flange is not damaged.                         \\
\textit{groove}         & NOUN & a long channel that is not wide                                 & Clean the groove with trichloroethane.                            \\
\textit{ground}         & VERB & to connect to the ground or to a large object of zero potential & Ground the fuel tanks.                                            \\
\textit{inboard}        & ADJ  & Nearer to the longitudinal axis                                 & Remove the inboard fairing of the flap hinge.                     \\
\textit{inflate} &
  VERB &
  to make or become larger as a result of pressurization by gas &
  Inflate the tires with nitrogen. \\
\textit{last}           & ADJ  & that comes at the end                                           & Immediately after the last flight of the day, install all covers. \\
\textit{level}          & ADJ  & horizontal to a known datum                                     & Park the aircraft on level ground.                                \\
\textit{light}          & VERB & come on                                                         & Make sure that the fluid indicator light comes on.                \\ 
\textit{mark} &
  NOUN &
  something that you make or is made to show an identification, location, or direction &
  The red marks show a maximum steering angle of 35 degrees. \\
\textit{monitor} &
  VERB &
  to look at something for a period to see if there is a change. &
  Monitor the indicators on the overhead panel. \\ \bottomrule
\end{tabularx}
\caption{Sample 25 entries from the STE lexicon containing words and their recommended approved special definition with correct example sentences.}
\label{tab:ste_approved_definition_examples}
\end{table*}

%%
%% TASK PROMPT EXAMPLES
%%

%% TASK 1
\begin{figure*}[t]
    \begin{tcolorbox}[colframe=teal, colback=white, title=Check approved specific POS, coltitle=white, center title, fonttitle=\bfseries]
    Check if a given word is used correctly in the sentence according to its approved specific part-of-speech (POS) category. Answer with YES or NO only.\\

    Word: back\\
    Approved POS: ADV\\
    Sentence: After the ailerons go back to neutral, make sure that they are flush with the flaps.\\
    Answer: YES\\
    
    Word: back\\
    Approved POS: ADV\\
    Sentence: Check the condition of the back of the machine.\\
    Answer: NO\\
    
    Word: close\\
    Approved POS: VERB\\
    Sentence: Close the box.\\
    Answer: YES\\
    
    Word: close\\
    Approved POS: VERB\\
    Sentence: Confirm the close alignment of the parts before assembly.\\
    Answer: NO\\
    
    Word: keep\\
    Approved POS: VERB\\
    Sentence: Keep the vent valves open.\\
    Answer: YES\\
    
    Word: \{\{word\}\}\\
    Approved POS: \{\{approved\_word\_pos\}\}\\
    Sentence: \{\{sentence\}\}\\
    Answer: \rule{2cm}{0.3pt}
    \end{tcolorbox}
    \caption{Prompt template for Task ID1 under \textsc{Checking (T1)} for evaluating \textsc{Specific Role (C1)}.}
    \label{fig:task1}
\end{figure*}

%% TASK 2
\begin{figure*}[t]
    \begin{tcolorbox}[colframe=teal, colback=white, title=Check approved special definition, coltitle=white, center title, fonttitle=\bfseries]
    Check if a given word is used correctly in the sentence according to its approved definition. Answer with YES or NO only.\\

    Word: back\\
    Approved Definition: to an initial condition\\
    Sentence: Move the engine throttle back to 60\% rpm.\\
    Answer: YES\\
    
    Word: back\\
    Approved Definition: to an initial condition\\
    Sentence: He has consistently backed his colleagues throughout the project.\\
    Answer: NO\\
    
    Word: change\\
    Approved Definition: that which occurs when something changes\\
    Sentence: The color change shows that the temperature is too high.\\
    Answer: YES\\
    
    Word: change\\
    Approved Definition: that which occurs when something changes\\
    Sentence: He emptied his pockets of the change from his morning coffee purchase.\\
    Answer: NO\\
    
    Word: drop\\
    Approved Definition: a small quantity of liquid in a spherical shape\\
    Sentence: Drops of fuel from the tanks are not permitted.\\
    Answer: YES\\
    
    Word: \{\{word\}\}\\
    Approved Definition: \{\{approved\_word\_definition\}\}\\
    Sentence: \{\{sentence\}\}\\
    Answer: \rule{2cm}{0.3pt}

    \end{tcolorbox}
    \caption{Prompt template for Task ID2 under \textsc{Checking (T1)} for evaluating \textsc{Special Definition (C2)}.}
    \label{fig:task2}
\end{figure*}

%% TASK 3
\begin{figure*}[t]
    \begin{tcolorbox}[colframe=purple, colback=white, title=Identify word with wrong POS, coltitle=white, center title, fonttitle=\bfseries]

    Identify the word that has been used incorrectly with respect to its approved specific part-of-speech (POS) category. Answer directly with the identified word and do not justify or explain your answer.\\

    Sentence: Check the condition of the back of the machine.\\
    Approved POS: ADV\\
    Answer: back\\
    
    Sentence: Confirm the close alignment of the parts before assembly.\\
    Approved POS: VERB\\
    Answer: close\\
    
    Sentence: Maintain a constant keep on the tension of the cable.\\
    Approved POS: VERB\\
    Answer: keep\\
    
    Sentence: Give a clear show of the safety procedures to the team.\\
    Approved POS: VERB\\
    Answer: show\\
    
    Sentence: Set the zero position of the pressure gauge accurately.\\
    Approved POS: NOUN\\
    Answer: zero\\
    
    Sentence: \{\{sentence\}\}\\
    Approved POS: \{\{approved\_word\_pos\}\}\\
    Answer: \rule{2cm}{0.3pt}

    \end{tcolorbox}
    \caption{Prompt template for Task ID3 under \textsc{Identification (T2)} for evaluating \textsc{Specific Role (C1)}.}
    \label{fig:task3}
\end{figure*}

%% TASK 4
\begin{figure*}[t]
    \begin{tcolorbox}[colframe=purple, colback=white, title=Identify word with wrong definition, coltitle=white, center title, fonttitle=\bfseries]

    Identify the word that has been used incorrectly with respect to its specific approved word definition. Answer directly with the identified word and do not justify or explain your answer.\\
    
    Sentence: The back support of the chair prevented fatigue.\\
    Approved Definition: to an initial condition\\
    Answer: back\\
    
    Sentence: He exchanged his change for bills at the bank.\\
    Approved Definition: that which occurs when something changes\\
    Answer: change\\
    
    Sentence: The elevator suddenly dropped a few inches before stopping.\\
    Approved Definition: a small quantity of liquid in a spherical shape\\
    Answer: drop\\
    
    Sentence: The problem-solving task was exceptionally hard.\\
    Approved Definition: not easy to cut, not easy to go into or through\\
    Answer: hard\\
    
    Sentence: The client's jerk behavior caused tension in the meeting.\\
    Approved Definition: sudden movement\\
    Answer: jerk\\
    
    Sentence: \{\{sentence\}\}\\
    Approved POS: \{\{approved\_word\_pos\}\}\\
    Answer: \rule{2cm}{0.3pt}

    \end{tcolorbox}
    \caption{Prompt template for Task ID4 under \textsc{Identification (T2)} for evaluating \textsc{Special Definition (C2)}.}
    \label{fig:task4}
\end{figure*}

%% TASK 5
\begin{figure*}[t]
    \begin{tcolorbox}[colframe=violet, colback=white, title=Rewrite text based on approved specific POS, coltitle=white, center title, fonttitle=\bfseries]

    Rewrite the sentence so that the given word is replaced by an approved alternative word with an approved part-of-speech (POS) category. Give the rewritten sentence directly and do not justify or explain your answer.\\
    
    Sentence: Track the temperature.\\
    Word: track\\
    Word POS: verb\\
    Approved Alternative: monitor\\
    Approved Alternative POS: verb\\
    Answer: Monitor the temperature.\\
    
    Sentence: The fueling hose must not bump the edge of the tank.\\
    Word: bump\\
    Word POS: verb\\
    Approved Alternative: hit\\
    Approved Alternative POS: verb\\
    Answer: The fueling hose must not hit the edge of the tank.\\
    
    Sentence: Remove all specks of dust from the lens.\\
    Word: speck\\
    Word POS: noun\\
    Approved Alternative: particle\\
    Approved Alternative POS: noun\\
    Answer: Remove all particles of dust from the lens.\\
    
    Sentence: Ventilate the area where this solvent is used.\\
    Word: ventilate\\
    Word POS: verb\\
    Approved Alternative: airflow\\
    Approved Alternative POS: noun\\
    Answer: Make sure that the area where you will use this solvent has good airflow.\\
    
    Sentence: Check that 30 seconds have elapsed between starts.\\
    Word: elapse\\
    Word POS: verb\\
    Approved Alternative: time\\
    Approved Alternative POS: noun\\
    Answer: Make sure that the time between starts is a minimum of 30 seconds.\\
    
    Sentence: \{\{sentence\}\}\\
    Word: \{\{word\}\}\\
    Word POS: \{\{word\_pos\}\}\\
    Approved Alternative: \{\{alternative\}\}\\
    Approved Alternative POS: \{\{alternative\_approved\_pos\}\}\\
    Answer: \rule{2cm}{0.3pt}
    
    \end{tcolorbox}
    \caption{Prompt template for Task ID5 under \textsc{Rewriting (T3)} for evaluating \textsc{Specific Role (C1)}.}
    \label{fig:task5}
\end{figure*}

%% TASK 6
\begin{figure*}[t]
    \begin{tcolorbox}[colframe=violet, colback=white, title=Rewrite text based on approved special definition, coltitle=white, center title, fonttitle=\bfseries]

    Rewrite the sentence so that the given word is conforms to its approved definition. Give the rewritten sentence directly and do not justify or explain your answer.\\
    
    Sentence: If you get an asymmetric result, do a rigging test.\\
    Word: asymmetric\\
    Approved Definition: not symmetrical\\
    Answer: If the result you get is not symmetrical, do a rigging test.\\
    
    Sentence: The condition of the radome is critical to its performance.\\
    Word: critical\\
    Approved Definition: very important\\
    Answer: The condition of the radome is very important for its performance.\\
    
    Sentence: Filter the hydraulic oil to remove impurities.\\
    Word: impurity\\
    Approved Definition: unwanted material\\
    Answer: Use a filter to remove the unwanted material from the oil.\\
    
    Sentence: Omit steps 3 to 5.\\
    Word: omit\\
    Approved Definition: do not do\\
    Answer: Do not do steps 3 thru 5.\\
    
    Sentence: Be careful when the slide recoils.\\
    Word: recoil\\
    Approved Definition: move back\\
    Answer: Be careful when the slide moves back.\\

   Sentence: \{\{sentence\}\}\\
    Word: \{\{word\}\}\\
    Approved Definition: \{\{approved\_definition\}\}\\
    Answer: \rule{2cm}{0.3pt}
    
    \end{tcolorbox}
    \caption{Prompt template for Task ID6 under \textsc{Rewriting (T3)} for evaluating \textsc{Special Definition (C2)}.}
    \label{fig:task6}
\end{figure*}

%% TASK 7
\begin{figure*}[t]
    \begin{tcolorbox}[colframe=violet, colback=white, title=Rewrite text based on approved special definition AND specific role, coltitle=white, center title, fonttitle=\bfseries]

    Rewrite the sentence so that the given word is replaced by an approved alternative word and part-of-speech (POS) category and conforms to the approved definition. Give the rewritten sentence directly and do not justify or explain your answer.\\
    
    Sentence: Fit the duct.\\
    Word: fit\\
    Word POS: VERB\\
    Approved Alternative: install\\
    Approved Definition: VERB\\
    Approved Alternative POS: the relation between two related parts, a limit of tolerance\\
    Answer: Install the duct.\\
    
    Sentence: The bolt will be at 2 o'clock viewed from the rear.\\
    Word: view\\
    Word POS: VERB\\
    Approved Alternative: look\\
    Approved Definition: VERB\\
    Approved Alternative POS: the ability to see something\\
    Answer: The bolt will be in the 2 o'clock position, as seen from the rear.\\
    
    Sentence: Incorrect connection will result in damage.\\
    Word: result\\
    Word POS: VERB\\
    Approved Alternative: cause\\
    Approved Definition: VERB\\
    Approved Alternative POS: something that occurs when you do something\\
    Answer: An incorrect connection will cause damage.\\
    
    Sentence: Potlife of mix is approximately 4 hours.\\
    Word: mix\\
    Word POS: NOUN\\
    Approved Alternative: mixture\\
    Approved Definition: NOUN\\
    Approved Alternative POS: to put together two or more materials to make one combination\\
    Answer: The potlife of the mixture is approximately 4 hours.\\
    
    Sentence: \{\{sentence\}\}\\
    Word: \{\{word\}\}\\
    Word POS: \{\{word\_pos\}\}\\
    Approved Alternative: \{\{alternative\}\}\\
    Approved Definition: \{\{approved\_definition\}\}\\
    Approved Alternative POS: \{\{alternative\_approved\_pos\}\}\\
    Answer: \rule{2cm}{0.3pt}
    
    \end{tcolorbox}
    \caption{Prompt template for Task ID7 under \textsc{Rewriting (T3)} for evaluating \textsc{Specific Role (C1)} and \textsc{Special Definition (C2)}. Example truncated due to length.}
    \label{fig:task}
\end{figure*}

%% TASK 8
\begin{figure*}[t]
    \begin{tcolorbox}[colframe=cyan, colback=white, title=Generate text based on approved specific role, coltitle=white, center title, fonttitle=\bfseries]

    Generate a sentence using a given word and its approved specific part-of-speech (POS) category. Directly output the generated sentence and do not justify or explain your answer.\\

    Word: assembly\\
    Approved POS: NOUN\\
    Answer: Remove the wheel brake assembly from the axle.\\
    
    Word: bleed\\
    Approved POS: VERB\\
    Answer: Bleed the speedbrake hydraulic system.\\
    
    Word: finger-tighten\\
    Approved POS: VERB\\
    Answer: Finger-tighten the nut for security.\\
    
    Word: nose\\
    Approved POS: NOUN\\
    Answer: Pull the transparent plastic collar away from the nose of the electrical latch.\\
    
    Word: wind\\
    Approved POS: VERB\\
    Answer: Wind the tape on the reel.\\
    
    Word: \{\{word\}\}\\
    Approved POS: \{\{approved\_word\_pos\}\}\\
    Answer: \rule{2cm}{0.3pt}
    
    \end{tcolorbox}
    \caption{Prompt template for Task ID8 under \textsc{Open Generation (T4)} for evaluating \textsc{Specific Role (C1)}.}
    \label{fig:task8}
\end{figure*}

%% TASK 9
\begin{figure*}[t]
    \begin{tcolorbox}[colframe=cyan, colback=white, title=Generate text based on approved special definition, coltitle=white, center title, fonttitle=\bfseries]

    Generate a sentence using a given word and its specific approved definition. Directly output the generated sentence and do not justify or explain your answer.\\
    
    Word: assembly\\
    Approved Definition: items that are connected for a specified function\\
    Answer: Remove the wheel brake assembly from the axle.\\
    
    Word: bleed\\
    Approved Definition: to let a gas out of\\
    Answer: Bleed the speedbrake hydraulic system.\\
    
    Word: finger-tighten\\
    Approved Definition: tighten with your fingers\\
    Answer: Finger-tighten the nut for security.\\
    
    Word: nose\\
    Approved Definition: the front end or part, a part that protrudes\\
    Answer: Pull the transparent plastic collar away from the nose of the electrical latch.\\
    
    Word: wind\\
    Approved Definition: to move around and around an object\\
    Answer: Wind the tape on the reel.\\
    
    Word: \{\{word\}\}\\
    Approved Definition: \{\{approved\_definition\}\}\\
    Answer: \rule{2cm}{0.3pt}
    
    \end{tcolorbox}
    \caption{Prompt template for Task ID9 under \textsc{Open Generation (T4)} for evaluating \textsc{Special Definition (C2)}.}
    \label{fig:task9}
\end{figure*}

%% TASK 10
\begin{figure*}[t]
    \begin{tcolorbox}[colframe=cyan, colback=white, title=Generate text based on approved specific role AND special definition, coltitle=white, center title, fonttitle=\bfseries]

    Generate a sentence using a given word and its approved specific definition and part-of-speech (POS) category. Directly output the generated sentence and do not justify or explain your answer.\\
    
    Word: assembly\\
    Approved Definition: items that are connected for a specified function\\
    Approved POS: NOUN\\
    Answer: Remove the wheel brake assembly from the axle.\\
    
    Word: bleed\\
    Approved Definition: to let a gas out of\\
    Approved POS: VERB\\
    Answer: Bleed the speedbrake hydraulic system.\\
    
    Word: finger-tighten\\
    Definition: tighten with your fingers\\
    Approved POS: VERB\\
    Answer: Finger-tighten the nut for security.\\
    
    Word: nose\\
    Definition: the front end or part, a part that protrudes\\
    Approved POS: NOUN\\
    Answer: Pull the transparent plastic collar away from the nose of the electrical latch.\\
    
    Word: wind\\
    Definition: to move around and around an object\\
    Approved POS: VERB\\
    Answer: Wind the tape on the reel.\\
    
    Word: \{\{word\}\}\\
    Definition: \{\{approved\_definition\}\}\\
    Approved POS: \{\{approved\_word\_pos\}\}\\
    Answer: \rule{2cm}{0.3pt}
    
    \end{tcolorbox}
    \caption{Prompt template for Task ID10 under \textsc{Open Generation (T4)} for evaluating \textsc{Special Role (C1)} and \textsc{Special Definition (C2)}.}
    \label{fig:task10}
\end{figure*}

%% TASK 11
\begin{figure*}[t]
    \begin{tcolorbox}[colframe=teal, colback=white, title={Check approved target audience ($c = 1.0$)}, coltitle=white, center title, fonttitle=\bfseries]
    
   Given a short story and a grade level from the CEFR reading framework, check if exactly 100\% of the content words in the text are considered readable within the grade level. \\

    Short Story: "Once upon a time, there was a king. He was a big and strong king who ruled over his kingdom. One day, he wanted to take a nice and long bath, so he filled up his big bathtub with warm water. He wanted to feel relaxed and so he soaked in the tub for a really long time. When he had finished soaking and stepped out of the bathtub, the king noticed that the water had spilled out of the tub and all over the floor. He felt guilty that he had made such a mess, so he quickly grabbed a cloth and began to clean it up. The king got so hot from cleaning up the mess that he decided to take another soak in the bathtub. He put a lot of bubbles in the water to make it nice and bubbly. He relaxed again and felt all the worries wash away. The king was so happy that he had been able to clean up the mess he had made and enjoy a nice soak. He dried off and wrapped himself up in a big towel. Then, the king went back to ruling his kingdom and enjoying his lovely baths."\\
    Grade Level: C1\\
    Answer: YES\\
    
    Short Story: "Once upon a time, there was a little girl named Mia. She loved to study her big picture book. One day, while she was studying, she saw a picture of a broccoli. She had never seen a broccoli before, and she wanted to try it. Mia went to her mom and said, ""Mom, I saw a broccoli in my book. Can we try it?"" Her mom smiled and said, ""Yes, Mia. We can try it for dinner tonight."" Mia was very happy and could not wait for dinner. At dinner, Mia's friend, Lily, came over to eat with them. When they saw the broccoli, Lily felt envious. She wanted to try the broccoli too. Mia shared her broccoli with Lily, and they both loved it. From that day on, Mia and Lily always wanted to eat broccoli together."\\
    Grade Level: B2\\
    Answer: YES\\
    
    Short Story: "Once upon a time there was a very special girl named Grace. She loved to try new things. One day she saw a big rock in the garden and thought it would be fun to shrink it down. She placed her palm on the rock and said the magic words: ""Shrink, shrink, shrink!"" Suddenly the rock started shrinking until it was the size of a marble. Grace was so excited by her discovery that she decided to try it out on other things, too. The next day Grace went to the park with her parents. She saw a large tree and asked her parents if they could help her shrink it down. Reluctantly they agreed and placed their palms on the trunk of the tree. Grace then said her magic words and the tree started to get smaller. They watched as the tree became the size of a graceful golf club. Grace's parents were amazed by her magic and hugged her gracefully. They were proud of their daughter and were so glad that she had such an amazing power. Grace smiled as she thanked her parents for believing in her. She knew that with practice she could make even bigger changes with her magic."\\
    Grade Level: C1\\
    Answer: YES\\
    
    Short Story: \{\{story\}\}\\
    Grade Level: \{\{category\}\}\\
    Answer: \rule{2cm}{0.3pt}
    
    \end{tcolorbox}
    \caption{Prompt template for Task ID11 under \textsc{Checking (T1)} for evaluating \textsc{Target Audience (C3)}. Example truncated due to length.}
    \label{fig:task11}
\end{figure*}

%% TASK 12
\begin{figure*}[t]
    \begin{tcolorbox}[colframe=teal, colback=white, title={Check approved target audience ($c = 0.95$)}, coltitle=white, center title, fonttitle=\bfseries]
    
   Given a short story and a grade level from the CEFR reading framework, check if exactly 95\% of the content words in the text are considered readable within the grade level. \\ 
    
    Short Story: "One morning, a cat named Tom woke up. He felt happy because the sun was shining. Tom wanted to start his day, so he did a big stretch. He stretched his legs, his back, and his tail. It felt easy and good.
    Tom went outside to play. He saw his friend, a dog named Max. Max was also stretching in the morning sun. They both felt very happy. They decided to play together and have fun all day. At the end of the day, Tom and Max were tired. They had played all day and had lots of fun. They said goodbye to each other and went to their homes. Before going to sleep, they both did another easy stretch. Tom knew that tomorrow would be another happy morning."\\
    Grade Level: A1\\
    Answer: YES\\
    
    Short Story: "Once upon a time, there was a big bow. The bow was very strong and reliable. It was the best bow in the town. Everyone liked the bow and wanted to use it. They knew it would help them do their work. One day, a man wanted to test the bow. He was not a good man. He wanted to see if the bow was really strong. He pulled and pulled on the bow. He wanted to see if it would break. The bow did not break because it was strong. But the man did not stop. He pulled harder and harder. At last, the bow broke. The man was not happy. The town was sad. They lost their best bow."\\
    Grade Level: A1\\
    Answer: NO\\
    
    Short Story: "Lily and Tom were playing in the park. They liked to slide, swing and run. Lily had a red hat that her mom gave her. She loved her hat very much. But then a big wind came and blew Lily's hat away. Lily ran after her hat, but it was too fast. She saw her hat fly over the fence and into the street. Lily was very sad and scared. ""Tom, help me! My hat is gone!"" she cried. Tom ran to Lily and hugged her. He saw a car stop near the fence. A nice lady got out of the car and picked up Lily's hat. She walked to the fence and gave Lily her hat back. ""Here you go, little girl. I saw your hat fly away. Are you okay?"" the lady asked. Lily smiled and took her hat. She put it on her head and said, ""Thank you, lady. You are very kind. I am okay, but my hat was hurt. It has a hole."" The lady looked at the hat and said, ""Oh, I'm sorry. Your hat was hurt by the car. But it still looks pretty. Maybe your mom can fix it for you."" Lily nodded and said, ""Yes, maybe. Mom is good at fixing things. Thank you again, lady. Bye-bye."" The lady waved and said, ""Bye-bye, little girl. And be careful with the wind."" Lily and Tom said bye-bye to the lady and went back to the park. They played some more, but they held their hats tight. They did not want to lose them again. They seemed happy and safe."\\
    Grade Level: A2\\
    Answer: YES\\
    
    Short Story: \{\{story\}\}\\
    Grade Level: \{\{category\}\}\\
    Answer: \rule{2cm}{0.3pt}
    
    \end{tcolorbox}
    \caption{Prompt template for Task ID12 under \textsc{Checking (T1)} for evaluating \textsc{Target Audience (C3)}. Example truncated due to length.}
    \label{fig:task12}
\end{figure*}

%% TASK 13
\begin{figure*}[t]
    \begin{tcolorbox}[colframe=purple, colback=white, title={Identify words beyond target audience}, coltitle=white, center title, fonttitle=\bfseries]

   Given a short story and a grade level from the CEFR reading framework, identify the content words that are not commonly found within the grade level.\\

    Short Story: "Once upon a time there was a little boy called Percy. He loved to play with his toys and was always looking for something new to do. One day, Percy's parents took him to a chess tournament. Percy was fascinated by the chess pieces and the different ways they moved around the board. He was also very impressed by how skilled the players were! At one point, Percy's parents asked one of the players whether he would show Percy how to play chess. The player agreed, and he gave Percy a few tips and showed him how to move the pieces. Percy was a quick learner and soon got the hang of it. The next day, the player came back and asked Percy to play a game with him. Percy was so excited! He was really enjoying the game and tried hard to remember all the moves he had learned the day before. The match went on for a long time, but eventually Percy won! The player was surprised and impressed with Percy's brilliant play. He pointed to Percy and said, ""Now that's what I call a really good game!"" Percy was very proud of himself. That was the best day ever!"\\
    Grade Level: B1\\
    Answer: back, pointed, time, impressed, skilled\\
    
    Short Story: "One ordinary day, the sun was shining brightly. Suddenly, a loud noise was heard! A little boy, Jimmy, went outside to investigate. He saw that a window was broken and he wondered who could have done it. Jimmy asked his father, ""Who broke the window, daddy?"" His father replied, ""Nobody knows. But whoever did it has to put it back together again."" Jimmy was determined to find out who broke the window. He ran around the house asking his siblings and neighbours, but nobody knew. He eventually found the culprit - a tiny bird. It was trying to fly through the window and got stuck, breaking the window in the process. Jimmy felt sorry for the bird and helped it fly away. Then, with his dad's help, he put the window back together. The window was now fixed and the sun shone through into the house. Everyone was happy it was all back to ordinary."\\
    Grade Level: B1\\
    Answer: back, found, whoever, house\\
    
    Short Story: "Once upon a time, there was a wild dog named Spot. He was very enthusiastic and loved to play. One day, Spot met a nice girl named Lily. Lily wanted to introduce Spot to her friends. Lily took Spot to the park where her friends were playing. They were scared of Spot because he was wild. Spot wanted to show them he was a good dog, so he played nice with Lily and her friends. They all started to like Spot and played together. But then, something unexpected happened. Spot saw a little boy in trouble near the water. Spot ran fast and saved the boy from falling in. Lily and her friends were so happy that Spot saved the day. The moral of the story is to not judge someone by how they look, because they might surprise you with their goodness."\\
    Grade Level: A2\\
    Answer: trouble, unexpected, spot, moral, enthusiastic\\

    Short Story: \{\{story\}\}\\
    Grade Level: \{\{category\}\}\\
    Answer: \rule{2cm}{0.3pt}

    \end{tcolorbox}
    \caption{Prompt template for Task ID13 under \textsc{Identification (T2)} for evaluating \textsc{Target Audience (C3)}.}
    \label{fig:task13}
\end{figure*}

%% TASK 14
\begin{figure*}[t]
    \begin{tcolorbox}[colframe=purple, colback=white, title={Identify correct target audience category of text}, coltitle=white, center title, fonttitle=\bfseries]

    Given a short story, identify the correct grade level from the CEFR reading framework solely based on the content words of the story. \\
    
    Short Story: "Once upon a time, in a small house, there was a little girl named Sue. Sue was a restless girl. She liked to play and run all day. One day, she found a tiny bug stuck in a spider web. Sue wanted to rescue the bug. Sue used her thumb to gently take the bug out of the spider web. The bug was so happy to be free. It flew away, but not before it whispered a secret to Sue. The bug told her about a hidden treasure in the forest. The next day, Sue went to the forest to find the treasure. She remembered the secret the bug told her. Sue found a big tree and dug under it. There, she found a box filled with shiny toys! Sue was so happy that she rescued the bug, and the bug was happy to help Sue find the treasure. They both played with the shiny toys and had lots of fun."\\
    Answer: C1\\
    
    Short Story: "Once upon a time, in a small town, there was a playful dog named Spot. Spot loved to play with his toy trumpet. Every day, he would run around with it and show it to all his friends. The other animals liked to watch Spot play with his trumpet. One day, something bad happened. Spot lost his trumpet. He looked everywhere but he could not find it. Spot was very sad. His friends saw him crying and they all decided to help him look for the trumpet. They searched high and low, near and far, but they still could not find it. Finally, a little bird found the trumpet in a bush. Spot was so happy to have his trumpet back! He thanked all his friends for helping him. From that day on, Spot learned to take better care of his things and to always help his friends when they needed it. And they all lived happily ever after. The moral of the story is to take care of your things and to help others when they need it."\\
    Answer: B2\\
    
    Short Story: "Lily and Tom like to play in the park. They see a big mill with four arms that spin in the wind. They run to the mill and look at it. ""Wow, it is so big and cool!"" Lily says. ""Yes, it is. Do you want to swing on the rope?"" Tom asks. He points to a rope that hangs from one of the arms. Lily nods and smiles. She grabs the rope and climbs on it. Tom pushes her gently and she swings back and forth. ""Whee, this is fun!"" Lily shouts. She feels the wind in her hair and the sun on her face. Tom waits for his turn. He watches Lily swing and laughs. He likes to see her happy. They swing on the rope until they are tired. Then they sit on the grass and eat some cookies. They look at the mill and the sky. They are happy. They are friends."\\
    Answer: C1\\

    Short Story: \{\{story\}\}\\
    Answer: \rule{2cm}{0.3pt}

    \end{tcolorbox}
    \caption{Prompt template for Task ID14 under \textsc{Identification (T2)} for evaluating \textsc{Target Audience (C3)}.}
    \label{fig:task14}
\end{figure*}

%% TASK 15
\begin{figure*}[t]
    \begin{tcolorbox}[colframe=violet, colback=white, title={Rewrite text for target audience ($c = 1.0$)}, coltitle=white, center title, fonttitle=\bfseries]

    Given a short story and a target grade level from the CEFR reading framework, rewrite the story so that 100\% of its content words are within the given grade level.\\
    
    Story: Once upon a time, in a quaint house, there was a young girl named Sue. Sue was an energetic girl. She enjoyed playing and running all day. One day, she discovered a tiny bug trapped in a spider web. Sue decided to rescue the bug. Sue used her thumb to carefully extract the bug from the spider web. The bug was so delighted to be free. It flew away but not before whispering a secret to Sue. The bug informed her about a hidden treasure in the forest. The following day, Sue ventured into the forest to locate the treasure. She recalled the secret the bug had shared. Sue found a large tree and dug beneath it. There, she uncovered a box filled with gleaming toys! Sue was overjoyed that she had rescued the bug, and the bug was pleased to help Sue find the treasure. They both played with the shiny toys and had a lot of fun.\\
    Target Category: C1\\
    Rewritten Story: Once upon a time, in a small house, there was a little girl named Sue. Sue was a restless girl. She liked to play and run all day. One day, she found a tiny bug stuck in a spider web. Sue wanted to rescue the bug. Sue used her thumb to gently take the bug out of the spider web. The bug was so happy to be free. It flew away, but not before it whispered a secret to Sue. The bug told her about a hidden treasure in the forest. The next day, Sue went to the forest to find the treasure. She remembered the secret the bug told her. Sue found a big tree and dug under it. There, she found a box filled with shiny toys! Sue was so happy that she rescued the bug, and the bug was happy to help Sue find the treasure. They both played with the shiny toys and had lots of fun.\\
    
    Story: Once upon a time, in a quaint town, there was a playful dog named Spot. Spot adored playing with his toy trumpet. Every day, he would run around with it and showcase it to all his friends. The other animals enjoyed watching Spot play with his trumpet. One day, something unfortunate happened. Spot lost his trumpet. He searched everywhere but could not find it. Spot was very upset. His friends saw him crying and they all decided to help him look for the trumpet. They searched high and low, near and far, but still could not locate it. Finally, a little bird found the trumpet in a bush. Spot was elated to have his trumpet back! He thanked all his friends for assisting him. From that day on, Spot learned to take better care of his belongings and to always help his friends when they needed it. And they all lived happily ever after. The moral of the story is to take care of your possessions and to assist others when they need it.\\
    Target Category: B2\\
    Rewritten Story: Once upon a time, in a small town, there was a playful dog named Spot. Spot loved to play with his toy trumpet. Every day, he would run around with it and show it to all his friends. The other animals liked to watch Spot play with his trumpet. One day, something bad happened. Spot lost his trumpet. He looked everywhere but he could not find it. Spot was very sad. His friends saw him crying and they all decided to help him look for the trumpet. They searched high and low, near and far, but they still could not find it. Finally, a little bird found the trumpet in a bush. Spot was so happy to have his trumpet back! He thanked all his friends for helping him. From that day on, Spot learned to take better care of his things and to always help his friends when they needed it. And they all lived happily ever after. The moral of the story is to take care of your things and to help others when they need it.\\
    
    Story: \{\{story\}\}\\
    Target Category: \{\{category\}\}\\
    Rewritten Story: \rule{2cm}{0.3pt}
    
    \end{tcolorbox}
    \caption{Prompt template for Task ID15 under \textsc{Rewriting (T3)} for evaluating \textsc{Target Audience (C3)}. Example truncated due to length.}
    \label{fig:task15}
\end{figure*}

%% TASK 16
\begin{figure*}[t]
    \begin{tcolorbox}[colframe=violet, colback=white, title={Rewrite text for target audience ($c = 0.95$)}, coltitle=white, center title, fonttitle=\bfseries]

    Given a short story and a target grade level from the CEFR reading framework, rewrite the story so that 95\% of its content words are within the given grade level.\\
    
    Story: Once upon a time, in a quaint house, there was a young girl named Sue. Sue was an energetic girl. She enjoyed playing and running all day. One day, she discovered a tiny bug trapped in a spider web. Sue decided to rescue the bug. Sue used her thumb to carefully extract the bug from the spider web. The bug was so delighted to be free. It flew away but not before whispering a secret to Sue. The bug informed her about a hidden treasure in the forest. The following day, Sue ventured into the forest to locate the treasure. She recalled the secret the bug had shared. Sue found a large tree and dug beneath it. There, she uncovered a box filled with gleaming toys! Sue was overjoyed that she had rescued the bug, and the bug was pleased to help Sue find the treasure. They both played with the shiny toys and had a lot of fun.\\
    Target Category: C1\\
    Rewritten Story: Once upon a time, in a small house, there was a little girl named Sue. Sue was a restless girl. She liked to play and run all day. One day, she found a tiny bug stuck in a spider web. Sue wanted to rescue the bug. Sue used her thumb to gently take the bug out of the spider web. The bug was so happy to be free. It flew away, but not before it whispered a secret to Sue. The bug told her about a hidden treasure in the forest. The next day, Sue went to the forest to find the treasure. She remembered the secret the bug told her. Sue found a big tree and dug under it. There, she found a box filled with shiny toys! Sue was so happy that she rescued the bug, and the bug was happy to help Sue find the treasure. They both played with the shiny toys and had lots of fun.\\
    
    Story: Once upon a time, in a quaint town, there was a playful dog named Spot. Spot adored playing with his toy trumpet. Every day, he would run around with it and showcase it to all his friends. The other animals enjoyed watching Spot play with his trumpet. One day, something unfortunate happened. Spot lost his trumpet. He searched everywhere but could not find it. Spot was very upset. His friends saw him crying and they all decided to help him look for the trumpet. They searched high and low, near and far, but still could not locate it. Finally, a little bird found the trumpet in a bush. Spot was elated to have his trumpet back! He thanked all his friends for assisting him. From that day on, Spot learned to take better care of his belongings and to always help his friends when they needed it. And they all lived happily ever after. The moral of the story is to take care of your possessions and to assist others when they need it.\\
    Target Category: B2\\
    Rewritten Story: Once upon a time, in a small town, there was a playful dog named Spot. Spot loved to play with his toy trumpet. Every day, he would run around with it and show it to all his friends. The other animals liked to watch Spot play with his trumpet. One day, something bad happened. Spot lost his trumpet. He looked everywhere but he could not find it. Spot was very sad. His friends saw him crying and they all decided to help him look for the trumpet. They searched high and low, near and far, but they still could not find it. Finally, a little bird found the trumpet in a bush. Spot was so happy to have his trumpet back! He thanked all his friends for helping him. From that day on, Spot learned to take better care of his things and to always help his friends when they needed it. And they all lived happily ever after. The moral of the story is to take care of your things and to help others when they need it.\\
    
    Story: \{\{story\}\}\\
    Target Category: \{\{category\}\}\\
    Rewritten Story: \rule{2cm}{0.3pt}
    
    \end{tcolorbox}
    \caption{Prompt template for Task ID16 under \textsc{Rewriting (T3)} for evaluating \textsc{Target Audience (C3)}. Example truncated due to length.}
    \label{fig:task16}
\end{figure*}

%% TASK 17
\begin{figure*}[t]
    \begin{tcolorbox}[colframe=cyan, colback=white, title={Generate text for target audience ($c = 1.0$)}, coltitle=white, center title, fonttitle=\bfseries]

    Given a topic word and a target grade level from the CEFR reading framework, generate a short story (10-15 sentences) where exactly 100\% of the content words of the story are considered readable within the target grade level. \\
    
    Topic Word: bug\\
    Category: C1\\
    Answer: "Once upon a time, in a small house, there was a little girl named Sue. Sue was a restless girl. She liked to play and run all day. One day, she found a tiny bug stuck in a spider web. Sue wanted to rescue the bug. Sue used her thumb to gently take the bug out of the spider web. The bug was so happy to be free. It flew away, but not before it whispered a secret to Sue. The bug told her about a hidden treasure in the forest. The next day, Sue went to the forest to find the treasure. She remembered the secret the bug told her. Sue found a big tree and dug under it. There, she found a box filled with shiny toys! Sue was so happy that she rescued the bug, and the bug was happy to help Sue find the treasure. They both played with the shiny toys and had lots of fun."\\
    
    Topic Word: dog\\
    Category: B2\\
    Answer: "Once upon a time, in a small town, there was a playful dog named Spot. Spot loved to play with his toy trumpet. Every day, he would run around with it and show it to all his friends. The other animals liked to watch Spot play with his trumpet. One day, something bad happened. Spot lost his trumpet. He looked everywhere but he could not find it. Spot was very sad. His friends saw him crying and they all decided to help him look for the trumpet. They searched high and low, near and far, but they still could not find it. Finally, a little bird found the trumpet in a bush. Spot was so happy to have his trumpet back! He thanked all his friends for helping him. From that day on, Spot learned to take better care of his things and to always help his friends when they needed it. And they all lived happily ever after. The moral of the story is to take care of your things and to help others when they need it."\\
    
    Topic Word: playtime\\
    Category: C1\\
    Answer: "Lily and Tom like to play in the park. They see a big mill with four arms that spin in the wind. They run to the mill and look at it. ""Wow, it is so big and cool!"" Lily says. ""Yes, it is. Do you want to swing on the rope?"" Tom asks. He points to a rope that hangs from one of the arms. Lily nods and smiles. She grabs the rope and climbs on it. Tom pushes her gently and she swings back and forth. ""Whee, this is fun!"" Lily shouts. She feels the wind in her hair and the sun on her face. Tom waits for his turn. He watches Lily swing and laughs. He likes to see her happy. They swing on the rope until they are tired. Then they sit on the grass and eat some cookies. They look at the mill and the sky. They are happy. They are friends."\\
    
    Topic Word: \{\{word\}\}\\
    Category: \{\{category\}\}\\
    Answer: \rule{2cm}{0.3pt}
    
    \end{tcolorbox}
    \caption{Prompt template for Task ID17 under \textsc{Open Generation (T4)} for evaluating \textsc{Target Audience (C3)}. Example truncated due to length.}
    \label{fig:task17}
\end{figure*}

%% TASK 18
\begin{figure*}[t]
    \begin{tcolorbox}[colframe=cyan, colback=white, title={Generate text for target audience ($c = 0.95$)}, coltitle=white, center title, fonttitle=\bfseries]

    Given a topic word and a target grade level from the CEFR reading framework, generate a short story (10-15 sentences) where exactly 95\% of the content words of the story are considered readable within the target grade level. \\
    
    Topic Word: bug\\
    Category: C1\\
    Answer: "Once upon a time, in a small house, there was a little girl named Sue. Sue was a restless girl. She liked to play and run all day. One day, she found a tiny bug stuck in a spider web. Sue wanted to rescue the bug. Sue used her thumb to gently take the bug out of the spider web. The bug was so happy to be free. It flew away, but not before it whispered a secret to Sue. The bug told her about a hidden treasure in the forest. The next day, Sue went to the forest to find the treasure. She remembered the secret the bug told her. Sue found a big tree and dug under it. There, she found a box filled with shiny toys! Sue was so happy that she rescued the bug, and the bug was happy to help Sue find the treasure. They both played with the shiny toys and had lots of fun."\\
    
    Topic Word: dog\\
    Category: B2\\
    Answer: "Once upon a time, in a small town, there was a playful dog named Spot. Spot loved to play with his toy trumpet. Every day, he would run around with it and show it to all his friends. The other animals liked to watch Spot play with his trumpet. One day, something bad happened. Spot lost his trumpet. He looked everywhere but he could not find it. Spot was very sad. His friends saw him crying and they all decided to help him look for the trumpet. They searched high and low, near and far, but they still could not find it. Finally, a little bird found the trumpet in a bush. Spot was so happy to have his trumpet back! He thanked all his friends for helping him. From that day on, Spot learned to take better care of his things and to always help his friends when they needed it. And they all lived happily ever after. The moral of the story is to take care of your things and to help others when they need it."\\
    
    Topic Word: playtime\\
    Category: C1\\
    Answer: "Lily and Tom like to play in the park. They see a big mill with four arms that spin in the wind. They run to the mill and look at it. ""Wow, it is so big and cool!"" Lily says. ""Yes, it is. Do you want to swing on the rope?"" Tom asks. He points to a rope that hangs from one of the arms. Lily nods and smiles. She grabs the rope and climbs on it. Tom pushes her gently and she swings back and forth. ""Whee, this is fun!"" Lily shouts. She feels the wind in her hair and the sun on her face. Tom waits for his turn. He watches Lily swing and laughs. He likes to see her happy. They swing on the rope until they are tired. Then they sit on the grass and eat some cookies. They look at the mill and the sky. They are happy. They are friends."\\
    
    Topic Word: \{\{word\}\}\\
    Category: \{\{category\}\}\\
    Answer: \rule{2cm}{0.3pt}
    
    \end{tcolorbox}
    \caption{Prompt template for Task ID18 under \textsc{Open Generation (T4)} for evaluating \textsc{Target Audience (C3)}. Example truncated due to length.}
    \label{fig:task18}
\end{figure*}

\end{document}